\documentclass[letterpaper, 10 pt, journal, twoside]{IEEEtran}

\newcommand{\mbs}[1]{\ensuremath{\boldsymbol{#1}}}
\DeclareMathAlphabet{\mbf}{OT1}{ptm}{b}{n}
\newcommand{\mc}[1]{\ensuremath{\mathcal{#1}}}

\usepackage{xcolor}
\usepackage{amsmath}
\usepackage{amssymb}
\usepackage{mathtools}
\usepackage[backend=bibtex,bibstyle=ieee,citestyle=numeric-comp,doi=false,isbn=false]{biblatex}
\usepackage{xspace}
\usepackage{multirow}
\usepackage{booktabs}
\usepackage{placeins}

\newcommand{\OlsonMaxMixPaper}{{Olson \emph{et al.}}}
\newcommand{\PfeiferMSMPaper}{{Pfeifer \emph{et al.}}}
\newcommand{\RosenRisePaper}{{Rosen \emph{et al.}}}

\newcommand{\trans}{{\ensuremath{\mathsf{T}}}}

\newcommand{\mbfhat}[1]{\ensuremath{\hat{\mbf{#1}}}}
\newcommand{\mbfcheck}[1]{\ensuremath{\check{\mbf{#1}}}}
\DeclareMathOperator*{\argmin}{arg\,min}
\newcommand{\inv}{^{-1}}
\newcommand{\rnums}{\mathbb{R}}
\newcommand{\norm}[1]{\ensuremath{\left\Vert#1\right\Vert}}
\newcommand{\ErrJacSolver}{\mbf{J}}

\newcommand{\del}{\partial}

\newcommand{\Hessdx}[1]{\frac{\del^2 #1}{\del \mbf{x} \del \mbf{x}^\trans}}
\newcommand{\Jac}[2]{\frac{\del #1}{\del #2}}
\newcommand{\Jacdx}[1]{\frac{\del #1}{\del \mbf{x}}}
\newcommand{\JacdxT}[1]{\frac{\del #1}{\del \mbf{x}^\trans}}

\newcommand{\ErrDim}{{n_\textrm{e}}}

\DeclareMathOperator*{\argmax}{arg\,max}

\newcommand{\GmmNormConst}[1]{{\gamma_{\text{#1}}}}

\newcommand{\StateDim}{{n_\textrm{x}}}
\newcommand{\NumFactors}{{n_f}}
\newcommand{\NumLandmarks}{{n_\ell}}
\newcommand{\eye}{\mbf{1}}

\newcommand{\FactorIdx}{i}

\newcommand{\NonGaussian}{{\textrm{non-gauss}}}
\newcommand{\SumMix}{{\textrm{SM}}}
\newcommand{\HessSumMix}{{\textrm{HSM}}}
\newcommand{\MaxSumMix}{{\textrm{MSM}}}
\newcommand{\MixCompIdx}{{k}}
\newcommand{\NumComp}{{n_\MixCompIdx}}
\newcommand{\kmax}{{\MixCompIdx^*}}

\newcommand{\SumComp}{\sum_{\MixCompIdx=1}^{\NumComp}}
\newcommand{\SingleFactor}{J}

\newcommand{\Gaussian}[1]{\mc{N}\left(#1\right)}

\newcommand{\SourcePoint}{{\mbf{m}}}
\newcommand{\TargetPoint}{{\mbf{p}}}
\newcommand{\SourcePointSet}{{\mc{M}}}
\newcommand{\TargetPointSet}{{\mc{T}}}
\newcommand{\SourcePointIdx}{{i}}
\newcommand{\TargetPointIdx}{{j}}
\newcommand{\NumPhysDims}{{N}}
\newcommand{\SourceFrame}{{s}}
\newcommand{\TargetFrame}{{t}}
\newcommand{\GeneralSE}{{SE (\NumPhysDims)}}

\newcommand{\navlie}{\texttt{navlie}\xspace}
\newcommand{\ceres}{\texttt{ceres}\xspace}
\newcommand{\dee}{\textrm{d}}
\newcounter{daggerfootnote}

\definecolor{myrevisioncolor}{RGB}{255, 0, 0}
\colorlet{ColorVariable}{black}
\newcommand{\revision}[1]{{\textcolor{ColorVariable}{#1}}}

\ifCLASSINFOpdf
\else
\fi

\begin{document}
%
\title{A Hessian for Gaussian Mixture Likelihoods in Nonlinear Least Squares}
%
%
%

\author{Vassili Korotkine, Mitchell Cohen, and James Richard Forbes$^{1}$%
  \thanks{Manuscript received: April 4, 2024; Revised June 18, 2024; Accepted July 14, 2024.}
  \thanks{This paper was recommended for publication by Javier Civera upon evaluation of the Associate Editor and Reviewers' comments.
    This work was supported by 
    the Natural Sciences and Engineering Research Council of Canada (NSERC)
    Alliance Grant in collaboration with Denso Corporation.
  } 
  \thanks{$^{1}$The authors are with the Department of Mechanical Engineering, McGill
    University, Montreal, QC H3A 0C3, Canada (e-mails:
    \texttt{\small{vassili.korotkine@mail.mcgill.ca}},
    \texttt{\small{mitchell.cohen3@mail.mcgill.ca}}, \texttt{\small{james.richard.forbes@mcgill.ca}}).}%
  \thanks{Digital Object Identifier (DOI): see top of this page.}
}
%
%


\markboth{IEEE Robotics and Automation Letters. Preprint Version. Accepted July, 2024}
{Korotkine \MakeLowercase{\textit{et al.}}: A Hessian For Gaussian Mixture Likelihoods in NLS}

%



\maketitle

\begin{abstract}
  This paper proposes a 
  novel
  Hessian approximation for Maximum a Posteriori estimation problems in robotics
  involving Gaussian mixture likelihoods.
  Previous approaches manipulate the Gaussian mixture likelihood into 
  a form that allows the problem to be represented as
  a nonlinear least squares (NLS) problem. The resulting Hessian approximation
  used within NLS solvers from these approaches neglects certain nonlinearities.
  The proposed Hessian approximation is 
  derived by setting the Hessians of the Gaussian mixture component errors to zero, which is the same starting 
  point as for the Gauss-Newton Hessian approximation for NLS, 
  and using the chain rule to account for additional nonlinearities. The proposed Hessian 
  approximation
  \revision{results in improved convergence}
  \revision{speed and uncertainty characterization}
  \revision{for simulated experiments, and similar performance
    to the state of the art
    on real-world experiments.}
  A method to maintain compatibility with existing solvers, such as \ceres, is also presented. 
  Accompanying software and supplementary material can be found at
  \url{https://github.com/decargroup/hessian_sum_mixtures}.
\end{abstract}
\begin{IEEEkeywords}
  Sensor Fusion, Localization, Optimization and Optimal Control, Probabilistic Inference,
  SLAM
\end{IEEEkeywords}

%
\IEEEpeerreviewmaketitle

\section{Introduction and Related Work}
\label{sec:intro}
\IEEEPARstart{E}{stimating}
the state of a system from noisy and
incomplete sensor data is a central task for autonomous systems. To describe inherent uncertainty in the
measurements and state, probabilistic tools are required. Gaussian
measurement likelihoods are commonly used in state estimation, allowing
estimation problems to be easily formulated as instances of nonlinear least squares (NLS)
optimization~\cite[\S4.3]{barfootStateEstimationRobotics2017}.
The NLS formulations for state estimation stem from considering the negative log-likelihood (NLL)
of the Gaussian, allowing a cancellation between the logarithm and the Gaussian exponent.
\revision{Furthermore, the structure of NLS problems allows for the use of
    efficient solvers such as \ceres~\cite{Agarwal_Ceres_Solver_2022}, whereas
    more general optimizers or non-parametric methods~\cite{doucet2001sequential} may be more
    computationally demanding. Therefore, the NLS structure, which for many problems
    arises from Gaussianity of the error model, is highly desirable. }
However, in practice, many sensor models are highly
non-Gaussian, as in underwater acoustic positioning~\cite{cheung2019non}, or
subsea Simultaneous Localization and Mapping (SLAM)~\cite{newman2003pure}.
\par
Even when the sensor
characteristics are well-modeled by Gaussian distributions, multimodal error
distributions arise in real-world situations involving ambiguity. Examples include
simultaneous localization and mapping (SLAM)
with unknown data associations~\cite{doherty2020probabilistic},
3D multi-object tracking~\cite{poschmann2020factor}, and loop-closure ambiguity
in pose-graph optimization~\cite{lee2013robust}.
In such cases, Gaussian measurement likelihoods are insufficient, and realizing
robust and safe autonomy requires more accurate modeling of the underlying distributions.
A popular parametric model used to represent multimodal distributions is the
Gaussian mixture model (GMM), composed of a weighted sum of Gaussian
components~\cite[\S2.3.9]{bishopPatternRecognitionMachine2006}. Extensions of
the \emph{incremental smoothing and mapping} framework iSAM2, such as
multi-hypothesis iSAM \cite{hsiao2019mh}, have been developed to represent
ambiguity in state estimation problems.
However, these approaches are
incompatible with standard NLS optimization methods and require a dedicated
solver. Hence, other research has focused on incorporating GMM measurement
likelihoods for use in standard NLS optimization frameworks.
In the GMM case, the NLL does not simplify since the logarithm cannot
cancel with the Gaussian exponents\revision{, \emph{meaning that GMMs are not
        directly compatible with NLS solvers.}}
A common method to overcome this is the
\emph{Max-Mixture}\revision{, introduced by \OlsonMaxMixPaper~\cite{olsonInferenceNetworksMixtures2013}}, which
approximates the summation over Gaussians with a maximum operator
, reducing the problem back to a single
Gaussian. This technique allows the problem to be cast into an
NLS form at the cost of introducing additional local
minima. \revision{\RosenRisePaper~\cite{rosenRobustIncrementalOnline2013}} introduce the extension of Robust Incremental least-Squares Estimation (RISE)
to non-Gaussian models, which
allows for error terms with arbitrary distributions to be cast into instances of
NLS minimization. The application of the framework
in~\cite{rosenRobustIncrementalOnline2013} to GMMs error terms is
presented in \revision{\PfeiferMSMPaper}~\cite{pfeiferAdvancingMixtureModels2021} and called the \emph{Sum-Mixture} method.
While the approach in~\cite{rosenRobustIncrementalOnline2013} works well for
its intended scope, its application to GMMs results in issues
when iterative optimization methods are used.
\par
\begin{figure}
    \centering
    \includegraphics[width=\columnwidth]{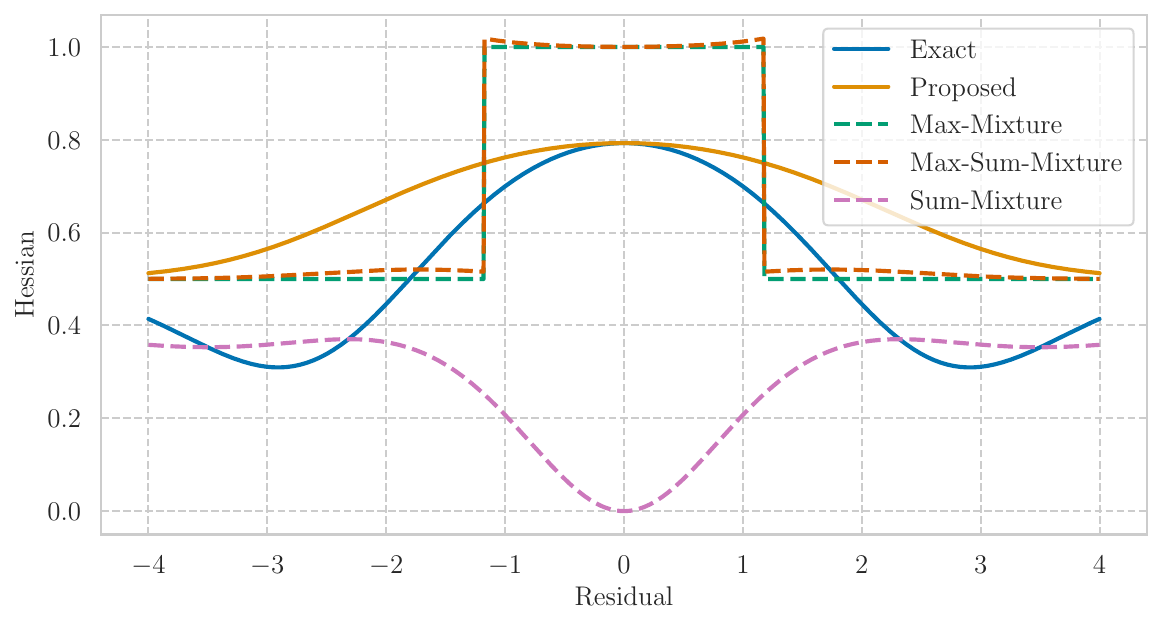}
    \vspace{-7mm}
    \caption{A comparison of the Hessian approximation for a Gaussian mixture factor
        consisting of two components, both centered around zero but with different covariances.
        For a scalar cost and state variable, the Hessian is also scalar. When substituted into a local optimization method,
        a better Hessian approximation results in better convergence.
        In particular, the proposed Hessian (orange) is much closer visually to
        the exact Hessian (green) than the Max-Mixture (blue) and Max-Sum-Mixture (yellow) approaches.
    }
    \label{fig:hessian_comparison}
\end{figure}
Common iterative methods
for solving NLS problems, such as Gauss-Newton, rely on an approximation of the
Hessian of the cost function.
\revision{
    Accuracy of the Hessian approximation manifests itself in convergence of the algorithm,
    since it is used to compute the
    step in NLS algorithms. Furthermore, the Hessian approximation can affect consistency, since
    the Laplace approximation~\cite[Sec. 4.4]{bishopPatternRecognitionMachine2006} uses the Hessian to
    provide a covariance on the estimate computed using the NLS algorithm.
}

When the Sum-Mixture approach is used in
conjunction with such optimization techniques, the resulting Hessian
approximation is inaccurate, leading to
degraded performance. This inaccuracy is pointed out by
\revision{\PfeiferMSMPaper}
in~\cite{pfeiferAdvancingMixtureModels2021}, but not explained theoretically.
\revision{\PfeiferMSMPaper}~\cite{pfeiferAdvancingMixtureModels2021} then propose an alternative to
the Sum-Mixture approach, termed the \emph{Max-Sum-Mixture} approach, which is a hybrid
between the Max-Mixture and Sum-Mixture approaches.
The dominant component of the mixture is factored out, creating a Max-Mixture error term, while
the remaining non-dominant factor is treated using the approach of \revision{\RosenRisePaper}~\cite{rosenRobustIncrementalOnline2013}.
The Hessian contribution for the non-dominant factor
is inaccurate in a similar fashion
as that of the Sum-Mixture, and improving
on this aspect is the focus of this paper.
\par
The contribution of this paper is a
novel Hessian approximation for NLS problems involving GMM terms. The proposed Hessian approximation is derived by
using the Gauss-Newton Hessian approximation
for the \emph{component errors} corresponding to each component of the GMM, and using the chain rule to
take into account the additional nonlinearities.
This approach is similar in spirit to how the robust loss nonlinearity is treated
when deriving the Triggs correction proposed in~\cite{triggs2000bundle} and used
in the \ceres~library~\cite{Agarwal_Ceres_Solver_2022}.
\revision{The proposed Hessian approximation has better mathematical properties in the sense that
    all mixture components contribute in the same way to the Hessian,
    without treating the dominant mixture component in a different framework than the rest,
    unlike other state-of-the-art approaches. Improvements to Hessian accuracy can
    therefore be expected in cases where multiple mixture components overlap.
}
\revision{Furthermore, }traditional NLS solvers do not allow the user to specify a separate Hessian, only
requiring an error definition and its corresponding Jacobian.
Therefore, this paper also proposes a method of defining an error and Jacobian that
uses the proposed Hessian while maintaining compatibility with the NLS
solver.
\par
\revision{
    The proposed Hessian approximation is showcased in
    Figure~\ref{fig:hessian_comparison} for a 1D example, where it is
    visually more accurate compared to state of the art.
    Performance of the proposed method is validated on a toy problem, a simulated point-set registration problem
    \cite{pfeiferAdvancingMixtureModels2021},~\cite{Jian2011} and on a SLAM problem with
    unknown data associations on the ``Lost in the Woods'' dataset~\cite{lostInTheWoodsDataset}.
}
The
implementation is open-sourced at \url{https://github.com/decargroup/hessian_sum_mixtures} and uses
the \navlie~NLS solver~\cite{cossette2023navlie}.
\par
The rest of this paper is organized as follows.
Newton's method and the Gauss-Newton Hessian approximation for NLS are presented in
Sec.~\ref{sec:theory}.
Sec.~\ref{sec:gaussian_mixtures} presents
Gaussian mixtures, the difficulties in applying NLS to Gaussian mixtures,
as well as a review of the previous approaches to solving this problem.
Sec.~\ref{sec:inaccuracy_msm} discusses the reasons for the Hessian inaccuracy of
existing methods, and Sec.~\ref{sec:proposed_approach} presents the
proposed approach.
Sec.~\ref{sec:nls_compatibility} describes how to maintain compability with traditional
NLS solvers while using the proposed Hessian.
Sec.~\ref{sec:results} presents simulation
and experimental results highlighting the benefits of the proposed approach.
\section{Theoretical Background}
\label{sec:theory}
Popular optimization methods used for state estimation are reviewed to motivate the improvement
of the Hessian approximation used. Furthermore, the link to NLL minimization
of Gaussian probabilities is reviewed,
while
Sec.~\ref{sec:gaussian_mixtures}
presents the extension to GMMs.
\subsection{Newton's Method}
\label{sec:newtons_method}
Newton's method is a local optimization method that iteratively minimizes a
quadratic approximation to the loss
function~\cite[\S4.3.1]{barfootStateEstimationRobotics2017}. Denoting the
optimization variable as $\mbf{x} \in \mathbb{R}^n$ and
the loss function to be minimized as
$J(\mbf{x})$, Newton's method
begins with an initial guess $\mbf{x}^{(0)}$, and iteratively updates the optimization variable using
\revision{$\mbf{x}^{(i+1)} =\mbf{x}^{(i)}+\alpha^{(i)} \Delta \mbf{x}^{(i)}$}
where $\alpha^{(i)} $ is a user-defined step size and $\Delta \mbf{x}^{(i)}$ is the
descent direction.
Setting the gradient of the
local quadratic approximation of the loss
at
$\mbf{x}^{(i)}$ to zero results in the descent direction being given by~\cite[\S2.2]{madsen2004methods}
\begin{align}
    \Hessdx{J(\mbf{x})} \Delta \mbf{x} & = -\frac{\del J(\mbf{x})}{\del \mbf{x}^\trans},
    \label{eq:newton_step}
\end{align}
where $\frac{\partial J (\mbf{x})}{\partial \mbf{x}} \revision{\in \mathbb{R}^{1\times n}}$
and $\Hessdx{J(\mbf{x})} \revision{\in \mathbb{R}^{n \times n}}$ are the
\revision{loss} Jacobian and Hessian respectively,
and are evaluated at the current iterate $\mbf{x}^{(i)}$.
The notation $\frac{\del J(\mbf{x})}{\del \mbf{x}^\trans} = \left(\frac{\del J(\mbf{x})}{\del \mbf{x}}\right)^\trans$
is used throughout this paper.
The iterate $\mbf{x}^{(i)}$
is updated and the process is iterated to convergence.
The main drawbacks of this method are the need to compute the Hessian, as well as the requirement
for the Hessian to be positive
definite. The Hessian is only guaranteed to be
positive semidefinite for convex losses~\cite[\S3.1.4]{boyd2004convex}, which are infrequent in robotics applications.
\subsection{Gauss-Newton Method}
\label{sec:gauss_newtons_method}
The Gauss-Newton method is an optimization approach applicable to problems that
have a NLS
structure. \revision{This structure is leveraged to form an approximate Hessian that is in turn used in the
    Newton step~\eqref{eq:newton_step}}.
The loss function is
written as the sum of squared error terms,
\begin{align}
    J(\mbf{x}) & =
    \sum_{\FactorIdx=1}^{\NumFactors} J_\FactorIdx(\mbf{x})=
    \sum_{\FactorIdx=1}^{\NumFactors}
    \frac{1}{2}
    \mbf{e}_\FactorIdx (\mbf{x})^\trans\mbf{e}_\FactorIdx (\mbf{x}),
    \label{eq:normalized_nll_least_squares}
\end{align}
where $J_\FactorIdx(\mbf{x})$ denotes a single summand,
$\NumFactors$ is the number of summands, \revision{and $\mbf{e}_i(\mbf{x}) \in \rnums^{n_{e, i}}$.}
The Jacobian and Hessian are computed using the chain rule,
\begin{align}
    \Jacdx{J(\mbf{x})} & =
    \sum_{\FactorIdx=1}^{\NumFactors}
    \Jac{J(\mbf{x})}{\mbf{e}_\FactorIdx} \Jac{\mbf{e}_\FactorIdx}{\mbf{x}}, \\
    \Hessdx{J(\mbf{x})}
                       & =
    \sum_{\FactorIdx=1}^{\NumFactors}
    \left(
    \Jac{\mbf{e}_\FactorIdx}{\mbf{x}}^\trans \Jac{\mbf{e}_\FactorIdx}{\mbf{x}}
    + \sum_{j=1}^{\ErrDim_\FactorIdx} e_{\FactorIdx, j} \Hessdx{e_{\FactorIdx,j}}
    \right),
    \label{eq:gauss_newton_hessian_full}
\end{align}
where the argument $\mbf{x}$ has been dropped in writing $\mbf{e}_i$.
The Gauss-Newton approximation sets the second term of~\eqref{eq:gauss_newton_hessian_full} to zero, such that
\begin{align}
    \Hessdx{J(\mbf{x})}
     & \approx \mbf{H}_{\text{GN}} =
    \sum_{\FactorIdx=1}^{\NumFactors}
    \Jac{\mbf{e}_\FactorIdx}{\mbf{x}}^\trans \Jac{\mbf{e}_\FactorIdx}{\mbf{x}},
    \label{eq:gauss_newton_hessian_approximation}
\end{align}
where $\mbf{H}_{\text{GN}}$ is the Hessian approximation made by Gauss-Newton.
For an error term $\mbf{e}_\FactorIdx (\mbf{x})$ affine in $\mbf{x}$, this approximation is exact, and the
algorithm converges in a single iteration.
\subsection{Maximum a Posteriori Estimation}
\label{sub:map_estimation}
\emph{Maximum a Posteriori} (MAP) estimation seeks to estimate the system state
$\mbf{x} \in \mathbb{R}^n$ by
maximizing the posterior probability~\cite[Sec.~3.1.2]{barfootStateEstimationRobotics2017},
which for Gaussian error models
results in the optimization problem
\begin{align}
    \mbfhat{x} & = \argmin_{\mbf{x}} \sum_{i=1}^{\NumFactors} -\log
    \Gaussian{\mbs{\eta}_\FactorIdx (\mbf{x}); \mbs{\mu}_\FactorIdx, \mbf{R}_\FactorIdx},
    \label{eq:nll_minimization_Gaussian}
\end{align}
where $\Gaussian{\mbs{\eta}_\FactorIdx; \mbs{\mu}_\FactorIdx, \mbf{R}_\FactorIdx}$
is a Gaussian probability density function (PDF)
in the error $\mbs{\eta}_\FactorIdx$ with
mean $\mbs{\mu}_\FactorIdx$ and covariance $\mbf{R}_\FactorIdx$.
The MAP estimation problem results in an NLS problem of the form~\eqref{eq:normalized_nll_least_squares}, with the
the normalized error $\mbf{e}_\FactorIdx(\mbf{x})$ obtained through the change of variables
\begin{align}
    \mbf{e}_\FactorIdx (\mbf{x} ) & =
    \sqrt{\mbf{R}_\FactorIdx\inv}(\mbs{\eta}_\FactorIdx (\mbf{x} )-\mbs{\mu}_\FactorIdx).
    \label{eq:change_of_variables}
\end{align}
For non-Gaussian error models,
the Gaussian in~\eqref{eq:nll_minimization_Gaussian}
is replaced by an arbitrary PDF $p(\mbs{\eta}_i)$ resulting in a more general summand
expression $J_i(\mbf{x})=-\log p(\mbs{\eta}_i)+c_i$ in~\eqref{eq:normalized_nll_least_squares}.
The constant $c_i$ absorbs extraneous normalization constants from $-\log p(\mbs{\eta}_i)$ that
do not affect the optimization.
\section{Gaussian Mixture Error Terms}
\label{sec:gaussian_mixtures}
A possible choice for the error model probability $p_i
    \left(\cdot \right)$ is a \emph{Gaussian mixture model}. Gaussian mixture models
are the sum of $n_k$ weighted Gaussian components, allowing for a wide
range of arbitrary probability distributions to be easily represented.
The expression for a single Gaussian mixture term, with the term index $\FactorIdx$
omitted for readability, is given by
\begin{align}
    \label{eq:Gaussian_mixture_error}
    p(\mbf{x}) & =
    \SumComp
    w_\MixCompIdx
    \Gaussian{\mbs{\eta}_\MixCompIdx(\mbf{x}); \mbs{\mu}_\MixCompIdx, \mbf{R}_\MixCompIdx},
\end{align}
where $w_k$ is the weight of the $k$'th Gaussian component. Note that here the sum is over components,
whereas in Sec.~\ref{sec:theory} the sums considered are over
problem error terms, of which only a single one is now considered.
To find an expression of the NLL for error models of the
form~\eqref{eq:Gaussian_mixture_error}, first let $\alpha_\MixCompIdx$ be
the weight normalized by covariance, written as
$\alpha_\MixCompIdx =
    w_\MixCompIdx
    \det(\mbf{R}_\MixCompIdx)^{-1/2}$.
Then, dropping state- and covariance-independent normalization constants
and using the change of variables~\eqref{eq:change_of_variables}
yields the expression for the corresponding summand given by
\begin{align}
    J_{\text{GMM}}(\mbf{x}) & =
    -\log p(\mbf{x}) + c        \\
                            & =
    -\log \SumComp
    \alpha_\MixCompIdx
    \exp \left(-\frac{1}{2}\mbf{e}_\MixCompIdx (\mbf{x} )^\trans \mbf{e}_\MixCompIdx (\mbf{x} )\right).
    \label{eq:nll_Gaussian_mixture}
\end{align}
The summand corresponding to a Gaussian mixture model NLL, given
by~\eqref{eq:nll_Gaussian_mixture}, is not of the form
of a standard NLS
problem in~\eqref{eq:normalized_nll_least_squares}. Unlike with Gaussian
error terms, an NLS problem cannot be directly obtained,
since the logarithm does not cancel the exponent. Hence, standard NLS optimization methods, such as Gauss-Newton, are not directly
applicable to problems with Gaussian mixture
error terms. Three existing approaches have been proposed to utilize Gaussian
mixture terms within the framework of NLS, as
discussed in the following sections.
%
\subsection{Max-Mixtures}
The Max-Mixture approach, first introduced
in \revision{\OlsonMaxMixPaper}~\cite{olsonInferenceNetworksMixtures2013}, replaces the
summation in~\eqref{eq:nll_Gaussian_mixture} with a $\max$ operator, such that
the NLL is given by
\begin{align}
    -\log p_{\max}(\mbf{x}) & =
    -\log \left(\max_k \alpha_\MixCompIdx \exp
    \left(-\frac{1}{2}\mbf{e}_\MixCompIdx (\mbf{x})^\trans \mbf{e}_\MixCompIdx (\mbf{x} )\right) \right).
\end{align}
The key is that the $\max$ operator and the logarithm may be swapped, allowing
the logarithm to be pushed inside the $\max$ operator to yield the Max-Mixture
error, given by
\begin{align}
    \label{eq:max_mixture_error}
    -\log p_{\max}(\mbf{x}) & =
    - \log \alpha_\MixCompIdx + \frac{1}{2}\mbf{e}_\kmax(\mbf{x} )^\trans \mbf{e}_\kmax (\mbf{x} ),
\end{align}
where
\begin{align}
    \kmax =
    \argmax_{\MixCompIdx}
    \alpha_\MixCompIdx \exp
    \left(-\frac{1}{2}\mbf{e}_\MixCompIdx (\mbf{x} )^\trans \mbf{e}_\MixCompIdx (\mbf{x} ) \right)
    \label{eq:dominant_index}
\end{align}
is the integer index of
the dominant mixture component at any given $\mbf{x}$. The Max-Mixture error
in~\eqref{eq:max_mixture_error} is in the standard NLS form, and methods such
as Gauss-Newton can be used to solve the problem.
\subsection{Sum-Mixtures}
A general method for treating non-Gaussian likelihoods in
NLS is proposed  in \revision{\RosenRisePaper}~\cite{rosenRobustIncrementalOnline2013}. For a general
non-Gaussian NLL
$-\log p_\NonGaussian(\mbf{x})$, the corresponding NLS error
$e_\NonGaussian(\mbf{x})$ is
derived by writing the relationship
\begin{align}
    J_\SumMix(\mbf{x}) & =
    -\log p_\NonGaussian(\mbf{x})+c \\
                       & =
    \frac{1}{2}
    \left(\sqrt{\log \GmmNormConst{\SumMix} -\log p_\NonGaussian(\mbf{x})}\right)
    ^2,
    \label{eq:general_sum_mix_nll}
\end{align}
where $\GmmNormConst{\SumMix}$ is a normalization constant chosen
such that $\GmmNormConst{\SumMix} \geq p_\NonGaussian(\mbf{x})$
to ensure that the square root argument is always positive.
In this case, the constant $c$ absorbs the constant offset that is added by the normalization constant.
The expression~\eqref{eq:general_sum_mix_nll} allows the error definition
\begin{align}
    e_\NonGaussian (\mbf{x}) & =
    \sqrt{\log \GmmNormConst{\SumMix} -\log p_\NonGaussian(\mbf{x})}.
    \label{eq:error_non_Gaussian}
\end{align}
The Sum-Mixture approach consists of applying this approach
to the Gaussian mixture case,
\begin{align}
    J_{\text{SM}} (\mbf{x}) & = \frac{1}{2}e_\SumMix^2 (\mbf{x}) \\
                            & =
    \frac{1}{2}
    \left(\sqrt{2}\sqrt{-\log \frac{1}{\GmmNormConst{\SumMix}}\SumComp \alpha_k \exp \left(-
        \frac{1}{2}\mbf{e}_\MixCompIdx^\trans
        \mbf{e}_\MixCompIdx\right)}\right)^2.
    \label{eq:sum_mixture_rosen_expression}
\end{align}
Ensuring positivity of the square root argument can be achieved
by setting the normalization constant to $\GmmNormConst{\SumMix}=\sum_{k=1}^\NumComp \alpha_k$~\cite{pfeifer2023adaptive}.
\subsection{Max-Sum-Mixtures}
The Max-Sum-Mixture is introduced by \revision{Pfeifer \emph{et al.}~\cite{pfeiferAdvancingMixtureModels2021}}, and is a hybrid between
the Max-Mixture and Sum-Mixture approaches.
The dominant component of the mixture is factored out such that the
corresponding summand is given by
\begin{align}
    \label{eq:max_sum_mixture_high_level_loss}
    J_{\text{MSM}} (\mbf{x})
     & =
    \frac{1}{2}\mbf{e}_{\kmax}^\trans \mbf{e}_\kmax \\
     & \quad -\log
    \sum_{k=1}^\NumComp \alpha_k
    \exp\left(-\frac{1}{2}\mbf{e}_k^\trans \mbf{e}_k+\frac{1}{2}\mbf{e}_{\kmax}^\trans \mbf{e}_\kmax\right).
    \nonumber
\end{align}
The first term  of~\eqref{eq:max_sum_mixture_high_level_loss}
is analogous to the Max-Mixture case. The second term is treated using the Sum-Mixture approach,
\begin{align}
    e_{2} (\mbf{x}) & =
    \sqrt{2}
    \sqrt{
        -\log
        \frac{1}{\GmmNormConst{\MaxSumMix}}
        \sum_{k=1}^\NumComp \frac{\alpha_k}{\alpha_\kmax}
        \exp(-\mbf{e}_k^\trans \mbf{e}_k+\mbf{e}_{\kmax}^\trans \mbf{e}_\kmax)
    },
\end{align}
where ${\GmmNormConst{\MaxSumMix}}$ is chosen as
${\GmmNormConst{\MaxSumMix}}=
    \NumComp \max \frac{\alpha_k}{\alpha_\kmax}+\delta$. The parameter $\delta > 0$
is a damping constant~\cite{pfeifer2023adaptive} that dampens
the influence of this nonlinear term in the optimization.
\section{Inaccuracy of the Non-Dominant Component Max-Sum-Mixture Hessian}
\label{sec:inaccuracy_msm}
NLS problems are defined by the choice of error. Different choices of error may be made, which in turn defines
the Gauss-Newton Hessian approximation. Because of the Gaussian noise assumption
in most robotics problems, the error choice is natural because the negative log-likelihood is quadratic.
However, since GMMs do not yield
a quadratic log-likelihood, there is no natural error choice.
The Sum-Mixture is one such error choice but yields an inaccurate Hessian approximation.
The inaccuracy is made clear by considering
a problem with a single Gaussian mixture error term of the
form~\eqref{eq:Gaussian_mixture_error}, having a single mode. The cost function
is given by $J (\mbf{x} ) = \frac{1}{2}\mbf{e} (\mbf{x} )^\trans \mbf{e} (\mbf{x} )$.
When the Sum-Mixture formulation is applied to this problem, the cost
function becomes
\begin{align}
    J (\mbf{x} ) & = \frac{1}{2}e_{\SumMix} (\mbf{x} )^2,
\end{align}
with $e_{\SumMix} (\mbf{x}) = \sqrt{\mbf{e}^\trans (\mbf{x} ) \mbf{e}
        (\mbf{x} )}$.
Even for a Gaussian likelihood problem that is easily solved by Gauss-Newton in its original formulation,
since the error $e_{\SumMix}(\mbf{x})$ is a scalar,
the Sum-Mixture
approach results in a rank one Hessian approximation~\eqref{eq:gauss_newton_hessian_approximation}
for the corresponding summand.
\revision{For state dimension $n>1$, the Hessian approximation is non-invertible
    and the Newton update~\eqref{eq:newton_step} fails.}
While this is not an issue within the original application
in~\cite{rosenRobustIncrementalOnline2013}, the Hessian approximation when applied to mixtures
is inaccurate as demonstrated with this base case.
\par
The fundamental reason for this inaccuracy is that higher-level
terms in the NLS Hessian~\eqref{eq:gauss_newton_hessian_full} are neglected.
Due to the added nonlinearity
$e_{\SumMix} (\mbf{x}) = \sqrt{\mbf{e}^\trans (\mbf{x} ) \mbf{e}
        (\mbf{x} )}$, terms get absorbed from the left term
of~\eqref{eq:gauss_newton_hessian_full} into the right one and end up neglected.
Even for well-behaved component errors,
the additional nonlinearity imposed by the LogSumExp expression inside the square root
in~\eqref{eq:sum_mixture_rosen_expression},
together with the square root itself, make it such that the Gauss-Newton approximation
is inaccurate, as demonstrated by the unimodal single-factor case.
The Max-Sum-Mixture approach
mostly addresses the shortcomings of the Sum-Mixture method by extracting the
dominant component.
\par
However, the non-dominant components in the second term of~\eqref{eq:max_sum_mixture_high_level_loss}
are still subjected to the same treatment. Their contribution to the overall Hessian is thus inaccurate, leading to the
discrepancies shown in Figure~\ref{fig:hessian_comparison}.

\section{Proposed Approach: Hessian-Sum-Mixture}
\label{sec:proposed_approach}
The proposed approach is termed the ``Hessian-Sum-Mixture'' (HSM) method and improves on previous formulations
by proposing a Hessian that takes into account the nonlinearity of the LogSumExp
expression in the GMM NLL~\eqref{eq:nll_Gaussian_mixture}, while keeping the Gauss-Newton approximation for each component.
The overall loss function is exactly the same, up to a constant offset,
as for the Sum-Mixture and Max-Sum-Mixture
approaches. However, the proposed Hessian derivation does not depend on defining
a Gauss-Newton error.
\par
The proposed Hessian-Sum-Mixture Hessian approximation is
derived by setting the second derivatives of the component errors to zero,
and using the chain rule to take into account the LogSumExp nonlinearity and
derive the cost Hessian.
\par
Formally, by writing the negative LogSumExp nonlinearity as
\begin{align}
    \rho(f_1, \dots, f_\NumComp) & =
    - \log \SumComp \alpha_\MixCompIdx\exp (-f_\MixCompIdx),
\end{align}
and setting the $f_\MixCompIdx$ components to the quadratic forms
$f_\MixCompIdx \left(\mbf{x} \right)=\frac{1}{2}\mbf{e}_\MixCompIdx \left(\mbf{x} \right)^\trans
    \mbf{e}_\MixCompIdx \left(\mbf{x} \right)$, the
negative Gaussian mixture log-likelihood may be rewritten as $-\log p(\mbf{x})=  \SingleFactor(\mbf{x})+c$ with
\begin{align}
    \SingleFactor(\mbf{x}) & = \rho(f_1(\mbf{e}_1(\mbf{x})), \dots, f_\MixCompIdx(\mbf{e}_\MixCompIdx(\mbf{x})), \dots
    f_\NumComp(\mbf{e}_\NumComp(\mbf{x}))).
    \label{eq:Gaussian_mixture_likelihood_thru_rho}
\end{align}
Using the Gauss-Newton Hessian approximation for the mixture components
yields
\begin{align}
    \Hessdx{f_k (\mbf{x})}\approx
    \frac{\del \mbf{e}_k}{\del \mbf{x}^\trans}
    \Jacdx{\mbf{e}_k}.
    \label{eq:f_k_gauss_newton_approximation}
\end{align}
Applying the chain rule to~\eqref{eq:Gaussian_mixture_likelihood_thru_rho}
and using the approximation~\eqref{eq:f_k_gauss_newton_approximation}
yields the factor Jacobian as
\begin{align}
    \Jac{J (\mbf{x})}{\mbf{x}} & =
    \sum_{k=1}^\NumComp
    \frac{\del \rho}{\del f_k}
    \Jacdx{f_k}, \label{eq:jac_Gaussian_mixture_first}
\end{align}
and the Hessian as
\begin{align}
    \Hessdx{\SingleFactor (\mbf{x})} & =
    \sum_{k=1}^\NumComp
    \left(
    \frac{\del \rho}{\del f_k}
    \Hessdx{f_k} +
    \frac{\del f_k}{\del \mbf{x}^\trans}
    \sum_{j=1}^\NumComp
    \frac{\del^2 \rho}{\del f_j f_k}
    \frac{\del f_j}{\del \mbf{x}}
    \right).
    \label{eq:hessian_Gaussian_mixture_first}
\end{align}
The partial derivatives of $\rho$ are given by
\begin{align}
    \frac{\del \rho}{\del f_k}
                                          & =
    \frac{\alpha_k \exp(-f_k)}{\sum_{i=1}^\NumComp\alpha_i\exp(-f_i)}
    \label{eq:first_rho_derivative}
    \\
    \frac{\del^2 \rho}{\del f_j \del f_k} & =
    \frac{
        -\delta_{jk} (\alpha_k \exp(-f_k))\sum_{i=1}^\NumComp \alpha_i\exp(-f_i)}
    {(\sum_{i=1}^\NumComp \alpha_i\exp(-f_i))^2} \nonumber \\
                                          & \quad +
    \frac{
        \alpha_k \alpha_j  \exp(-f_k)\exp(-f_j)}
    {(\sum_{i=1}^\NumComp \alpha_i\exp(-f_i))^2},
\end{align}
where $\delta_{jk}$ is the Kronecker delta
and the Jacobian of
$f_k \left( \mbf{x} \right) = \frac{1}{2}\mbf{e}_k{\left(\mbf{x}\right)}^\trans
    \mbf{e}_k{\left(\mbf{x}\right)}$
is given by
\begin{align}
    \Jacdx{f_k \left(\mbf{x} \right)} & =\mbf{e}_k
    {\left(\mbf{x} \right)}^\trans \Jacdx{\mbf{e}_k {\left(\mbf{x} \right)}}.
\end{align}
Using only the first of the terms in the summand of~\eqref{eq:hessian_Gaussian_mixture_first}
guarantees a positive definite Hessian approximation since $\frac{\del \rho}{\del f_k} $ is guaranteed to be positive.
The second term has no such guarantee. The final proposed Hessian approximation is
\begin{align}
    \Hessdx{\SingleFactor \left(\mbf{x} \right)} & \approx \mbf{H}_{\text{HSM}} =
    \sum_{k=1}^\NumComp
    \frac{\del \rho}{\del f_k}
    \Hessdx{f_k},
    \label{eq:hessian_Gaussian_mixture_our_approx}
\end{align}
where $\Hessdx{f_k}$ is approximated using~\eqref{eq:f_k_gauss_newton_approximation}.
Examining~\eqref{eq:first_rho_derivative} allows a somewhat intuitive
interpretation of~\eqref{eq:hessian_Gaussian_mixture_our_approx}. The
$\frac{\del \rho}{\del f_k}$ term is the relative strength of each component
at the evaluation point, while $\Hessdx{f_k}$ is the Hessian of that component.
The Hessian approximation $\mbf{H}_{\text{HSM}}$ in~\eqref{eq:hessian_Gaussian_mixture_our_approx}
corresponding to each summand is substituted directly into Newton's method
equation~\eqref{eq:newton_step}, instead of the Gauss-Newton
approximation in~\eqref{eq:gauss_newton_hessian_approximation}.
\par
The key difference with respect to Max-Sum-Mixture is that the dominant and non-dominant components are all
treated in the same manner. In the Max-Sum-Mixture the dominant component has a full-rank Hessian
contribution, while the non-dominant components have a rank one Hessian contribution
that is inaccurate as detailed in Sec.~\ref{sec:inaccuracy_msm}.
In situations where the components overlap significantly, the Hessian is thus more accurate in the proposed method.

\section{Nonlinear Least Squares Compatibility}
\label{sec:nls_compatibility}
Most NLS solvers do not support a separate Hessian definition such that the user
may only specify an error and error Jacobian, which then are input to
the GN approximation~\eqref{eq:gauss_newton_hessian_approximation}.
This may be circumvented by engineering an ``error'' and ``error Jacobian''
that result in the same descent direction as using Newton's method~\eqref{eq:newton_step}
with the HSM Hessian~\eqref{eq:hessian_Gaussian_mixture_our_approx}, while
maintaining a similar cost function value to the Sum-Mixture~\eqref{eq:general_sum_mix_nll}.
First, defining the following quantities
\begin{align}
    \mbf{e}_{\textrm{solver}, 1}^\trans       & =
    \begin{bmatrix}
        \sqrt{\frac{\del \rho}{\del f_1}}
        \mbf{e}_1^\trans & \dots &
        \sqrt{\frac{\del \rho}{\del f_K}} \mbf{e}_K^\trans
    \end{bmatrix},                    \\
    \ErrJacSolver_{\textrm{solver}, 1}^\trans & =
    \begin{bmatrix}
        \sqrt{\frac{\del \rho}{\del f_1}}
        \JacdxT{\mbf{e}_1} & \dots & \sqrt{\frac{\del \rho}{\del f_K}} \JacdxT{\mbf{e}_K}
    \end{bmatrix}, \label{eq:errorJacSolver}
\end{align}
and substituting into the GN Hessian approximation~\eqref{eq:gauss_newton_hessian_approximation} results
in exactly the Hessian-Sum-Mixture Hessian~\eqref{eq:hessian_Gaussian_mixture_first}
as well as the loss Jacobian~\eqref{eq:jac_Gaussian_mixture_first}.
However, the resulting cost function $\norm{\mbf{e}_{\textrm{solver}, 1}}_2^2$ is not of the same form
as~\eqref{eq:nll_Gaussian_mixture},
which can cause issues with methods that use the cost function to guide the optimization
such as Levenberg-Marquardt (LM)~\cite{madsen2004methods}.
To rectify this, an additional error term is required to make the loss into the same form
as~\eqref{eq:nll_Gaussian_mixture}.
The proposed error and Jacobian to be input into the solver are thus
\begin{align}
    \mbf{e}_{\textrm{solver}}^\trans       & =
    \begin{bmatrix}
        \mbf{e}_{\textrm{solver}, 1}^\trans & \sqrt{\revision{2(}\GmmNormConst{\HessSumMix} + \Delta J\revision{)}}
    \end{bmatrix},
    \label{eq:nls_compatibility_error}         \\
    \ErrJacSolver_{\textrm{solver}}^\trans & =
    \begin{bmatrix}
        \ErrJacSolver_{\textrm{solver, 1}}^\trans & 0
    \end{bmatrix},
\end{align}
where $\Delta J$ is the difference between the desired loss $J_{\textrm{GMM}}$~\eqref{eq:nll_Gaussian_mixture}
and the squared norm of $\mbf{e}_{\textrm{solver}, 1}$,
\begin{align}
    \Delta J = J_{\textrm{GMM}} - \revision{\frac{1}{2}}\norm{ \mbf{e}_{\textrm{solver}, 1}}_2^2,
    \label{eq:delta_J}
\end{align}
and $\GmmNormConst{\HessSumMix}$ is a normalization constant that ensures
positivity of the square root in~\eqref{eq:nls_compatibility_error}.
The evaluated \revision{cost $\frac{1}{2}\norm{\mbf{e}_{\textrm{solver}}}_2^2$} is thus
$\revision{\frac{1}{2}}\norm{\mbf{e}_{\textrm{solver}}}_2^2 = J_{\textrm{GMM}}+\GmmNormConst{\HessSumMix}$.
\revision{
Note that since
$\ErrJacSolver_{\textrm{solver}}$ is not the true Jacobian of the error $\mbf{e}_{\textrm{solver}}$,
caution must be taken if using automatic differentiation.
The normalization constant $\GmmNormConst{\HessSumMix}$ is taken to be the lower bound on
$\Delta J$ to ensure positivity of the square root argument.
By taking the logarithm and exponent of
the second term in~\eqref{eq:delta_J}, and using the properties of exponents,~\eqref{eq:delta_J}
becomes
\begin{align}
    \Delta J
        & =
    -\log \sum_{k=1}^\NumComp
    \left(
    \alpha_k
    \exp \left(
        S_k
        \right)
    \right), \\
    S_k & =
    \sum_{j=1}^\NumComp
    \frac{
        \alpha_j \exp\left(-f_j  \right)
    }{
        \sum_{i=1}^\NumComp \alpha_i \exp\left(-f_i \right)
    }
    \left(f_j-f_k\right),
\end{align}
with $f_k=\frac{1}{2}\mbf{e}_k^\trans \mbf{e}_k$. A sequence of algebraic manipulations
and using the fact that $t\exp(-t)\leq \exp(-1)$ yields
$S_k \leq
    \frac{1}{\alpha_k}
    \sum_{j=1}^\NumComp
    \alpha_j$, and thus
\begin{align}
    \Delta J \geq
    -    \log \sum_{k=1}^\NumComp
    \left(
    \alpha_k
    \exp \left(
        \frac{1}{\alpha_k}
        \sum_{j=1}^\NumComp \alpha_j
        \right)
    \right).
    \label{eq:delta_J_lower_bound}
\end{align}
The negative of the lower bound~\eqref{eq:delta_J_lower_bound} is used as $\GmmNormConst{\HessSumMix}$.
}

\section{Simulation and Experiments}
\label{sec:results}
To show the benefits of the proposed approach, the performance of the
Max-Mixture (MM), Max-Sum-Mixture (MSM), Sum-Mixture (SM), the Hessian Sum-Mixture (HSM),
are all
evaluated on two simulated experiments and one real dataset.
\revision{A benefit of HSM is in treating all of the components in the same
    way, unlike MM or MSM that privilege the dominant component. The simulated examples
    thus focus on cases where components overlap. }
The considered
simulation examples are similar to the ones provided
in~\cite{pfeiferAdvancingMixtureModels2021}, and the real-world experiment is a
SLAM problem with unknown data associations. In some of the tested examples, particularly
the 2D toy example case, the
Sum-Mixture approach fails when using standard Gauss-Newton due to a rank-deficient Hessian. To provide a fair comparison, Levenberg-Marquardt \cite{madsen2004methods}
(LM) is used instead of Gauss-Newton \revision{for all approaches}. LM adds a multiple of the
identity to the approximate Hessian, written as $\mbf{H}^\text{LM} = \mbf{H} +
    \lambda \mbf{I}$,  where $\lambda$ is a damping constant for the LM method,
$\mbf{H}$ is the original Hessian
approximation given by~\eqref{eq:gauss_newton_hessian_approximation} for the
Max-Mixture, Sum-Mixture, and Max-Sum-Mixture, and
by~\eqref{eq:hessian_Gaussian_mixture_our_approx} for the Hessian Sum Mixture
approach. This ensures that the Hessian approximation is always full rank. The LM
damping constant $\lambda$ is set as described in~\cite{madsen2004methods}.
\par
The metrics used to evaluate the quality of the obtained solution are
root-mean-squared error (RMSE), defined as
$\text{RMSE} =\sqrt{\frac{1}{N}\sum_{i=1}^N \norm{\mbfhat{x}_i \ominus \mbf{x}_i}_2^2}$,
where $\mbfhat{x}_i$ and $\mbf{x}_i$ are, respectively, the
estimated and ground truth states at timestep $i$. The $\ominus$
operator represents a generalized ``subtraction'' operator, required as
some examples considered involve states defined on Lie groups. Following the
definitions in~\cite{sola2021micro}, a left perturbation is used for all Lie group
operations
in this paper. Additionally, for states that involve rotations, the difference between the states is split into a rotational
and translational component, yielding a rotational RMSE (deg) and
a translational RMSE (m).
\par
Consistency, the ability of the estimator to characterize its error uncertainty, is measured using
average
normalized estimation error squared (ANEES)~\cite{Li2002}.
A perfectly consistent estimator has ANEES equal to one.
The main drawback of this metric is that it characterizes the posterior state belief
as Gaussian, which is never completely true in practice.
\par
Lastly, the average number of iterations \revision{and runtime} are used to characterize how quickly the estimator
converges to the solution. A smaller number of iterations indicates better descent directions
in~\eqref{eq:newton_step}, which is expected if a better Hessian approximation is used.
The solver exits once the step size of the LM algorithm is below $1 \times 10^{-8}$ or a cap of 200 iterations
is reached.
\subsection{Toy Example}
\label{sec:toy_example}
\begin{table}
    \centering
    \caption{Results of the toy example Monte Carlo experiment}
    \begin{tabular}{|*{6}{c|}}
        \hline
        \multirow{ 2}{*}{} Dims. & Method                   & RMSE              & Avg          & Time              & Succ.         \\
                                 &                          & (m)               & Iter.        & (s)               & Rate [\%]     \\
        \hline
        \multirow[c]{4}{*}{1D}   & MM                       & 4.07e-01          & \textbf{1.4} & \textbf{4.24e-03} & 24.1          \\
                                 & SM                       & 1.84e-02          & 26.1         & 4.12e-02          & 98.9          \\
                                 & MSM                      & \textbf{1.66e-02} & 18.6         & 3.10e-02          & \textbf{99.0} \\
                                 & \revision{HSM$^\dagger$} & 1.67e-02          & 8.8          & 1.50e-02          & \textbf{99.0} \\ \hline
        \multirow[c]{4}{*}{2D}   & MM                       & 7.16e-01          & \textbf{1.5} & \textbf{4.42e-03} & 30.4          \\
                                 & SM                       & 5.30e-02          & 27.9         & 4.61e-02          & 97.5          \\
                                 & MSM                      & \textbf{4.83e-02} & 12.9         & 2.31e-02          & \textbf{97.8} \\
                                 & \revision{HSM$^\dagger$} & 4.89e-02          & 9.1          & 1.57e-02          & \textbf{97.8} \\
        \hline
    \end{tabular}
    \label{tab:toy_results}
\end{table}
The first simulated example consists of optimizing a problem with a
Gaussian mixture model error term of the form
\begin{align}
    J(\mbf{x}) & =-\log \SumComp w_\MixCompIdx \Gaussian{\mbf{x};
        \mbfcheck{x}_\MixCompIdx, \mbf{R}_\MixCompIdx}.
\end{align}
The toy example experiments in this section use four mixture components, such that $\NumComp=4$.
To quantify the performance of all algorithms considered, Monte-Carlo
trials are run, where in each trial, random mixture parameters
are generated and run for many different initial conditions. Both 1D and 2D examples are
considered, where in the 1D case, the design variable is a scalar,  and in the
2D case, $\mbf{x} \in \mathbb{R}^2$. In all trials, the first mixture component is chosen with weight
$w_1 \sim \mathcal{U}(0.2, 0.8)$, where $\mathcal{U}(a, b)$ denotes the uniform
distribution over the interval $[a, b]$. The weights for the other mixture components
component are chosen as $w_k = (1-w_1)/(\NumComp-1)$. The mean for the first mixture
component is set to $\mbfcheck{x}_1 = \mbf{0}$, while each component of the
mean of the other mixture components is chosen as $\check{x}_2^i \sim \mathcal{U}(-2,
    2)$.
The covariance for the first component is set to $\mbf{R}_1 = \sigma_1^2
    \mbf{1}$, where $\sigma_1^2 \sim \mathcal{U}(0.4, 1)$.
The second component covariance is set to a multiple of the first
component such that $\mbf{R}_2 = m \mbf{R}_1$, where
$m \sim \mathcal{U}(4, 10)$.
For each generated set of mixture parameters, the optimization is run from 100
different initial points, chosen uniformly on a grid
on $[-4,4]$ for each axis in the problem dimension. Additionally, 1000 different
mixture parameters are generated, resulting in a total of 100, 000
Monte-Carlo trials for both the 1D example and 2D example.
The ground truth is determined by sampling on a grid combined with local optimization
to find the global optimum.
The result is
marked as \revision{successful} if it converges to the ground truth within a given
threshold, set to $0.01$ meters for all experiments. The results for both the 1D and 2D
Monte-Carlo experiments are shown in Table~\ref{tab:toy_results}. In all tables,
the proposed method is denoted by $^\dagger$.
The posterior in this case is extremely non-Gaussian, thus the ANEES is not reported.
All methods achieve similar error values, with Sum-Mixture presenting a small improvement over the other methods.
Note however that this is due to its interaction with LM, since for the 2D case particularly, the
Gauss-Newton version of Sum-Mixture results in a non-invertible Hessian approximation and the method does not work.
Nevertheless, the HSM approach still achieves
a lower number of iterations than SM or MSM.
The Max-Mixture approach achieves the fewest iterations but has a very low success rate.
\vspace{-0.1cm}
\subsection{Point-Set Registration}
\label{sec:point_set_registration}
\begin{table}
    \centering
    \caption{Results of the point-set registration experiment.}
    \begin{tabular}{|*{7}{c|}}
        \hline
        \multirow{ 2}{*}{}  Dims. & Method                   & RMSE          & RMSE          & ANEES         & Avg            & Time          \\
                                  &                          & (deg)         & (m)           &               & Iter.          & (s)           \\
        \hline
        \multirow[c]{4}{*}{2D}    & MM                       & 1.59          & 0.03          & 2.92          & \textbf{13.54} & \textbf{1.75} \\
                                  & SM                       & \textbf{1.15} & \textbf{0.02} & 0.05          & 55.05          & 9.50          \\
                                  & MSM                      & \textbf{1.15} & \textbf{0.02} & 1.61          & 25.79          & 4.27          \\
                                  & \revision{HSM$^\dagger$} & \textbf{1.15} & \textbf{0.02} & \textbf{1.44} & 22.56          & 3.66          \\ \hline
        \multirow[c]{4}{*}{3D}    & MM                       & 1.10          & 0.03          & 1.81          & \textbf{14.71} & \textbf{3.06} \\
                                  & SM                       & \textbf{0.95} & \textbf{0.02} & 0.03          & 71.08          & 20.53         \\
                                  & MSM                      & \textbf{0.95} & \textbf{0.02} & 1.34          & 22.81          & 6.02          \\
                                  & \revision{HSM$^\dagger$} & \textbf{0.95} & \textbf{0.02} & \textbf{1.26} & 21.24          & 5.51          \\
        \hline
    \end{tabular}

    \label{tab:psr_results}
\end{table}
The point-set registration example follows the structure of the example in~\cite{pfeiferAdvancingMixtureModels2021},
which itself builds on the formulation of~\cite{Jian2011}.
Given an uncertain source set of points $\mathcal{M} =
    \left\{\SourcePoint_\SourcePointIdx\right\}_{i = 1}^M$, as well as an uncertain reference
set of points $\TargetPointSet = \left\{\TargetPoint_\TargetPointIdx \right\}_{j=1}^{F}$
with $\SourcePoint_\SourcePointIdx, \TargetPoint_\TargetPointIdx \in
    \rnums^{\NumPhysDims}$,
the point-set registration task seeks to find the rigid-body transformation
$\mbf{T}_{\TargetFrame \SourceFrame}^\star \in \GeneralSE$ that best aligns
$\SourcePointSet$ to $\TargetPointSet$, where $N$ is either $2$ or $3$.
The transformation is found by forming
point-to-point residuals, $\mbf{e}_{ij}$,
\begin{align}
    \mbf{e}_{ij}(\mbf{p}_j, \mbf{m}_i , \mbf{C}_{ts}, \mbf{r}_t^{st}) & =
    \mbf{p}_j - \mbf{C}_{ts}\mbf{m}_i - \mbf{r}_t^{st},
\end{align}
where $\mbf{C}_{ts} \in SO(N)$ and $\mbf{r}_t^{st} \in \mathbb{R}^N$ are the
rotation and translation components of $\mbf{T}_{ts}$, respectively.
These errors are assumed Gaussian $\mbf{e}_{ij}\sim \Gaussian{\mbf{e}_{ij};\
        \mbf{0}, \mbf{R}_{ij}}$ with covariance
\begin{align}
    \mbf{R}_{ij}= \mbf{C}_{ts} \mbs{\Sigma}_m \mbf{C}_{ts}^\trans + \mbs{\Sigma}_f,
\end{align}
where $\mbs{\Sigma}_m$, $\mbs{\Sigma}_f$ are the noise covariances on
point measurements of points in $\SourcePointSet$ and $\TargetPointSet$, respectively.
Similarly to~\cite{pfeiferAdvancingMixtureModels2021}, to incorporate data
association ambiguity, the likelihood of a measurement $\mbf{m}_i$
is given by the Gaussian mixture
\begin{align}
    p(\mbf{m}_i | \mbf{C}_{ts}, \mbf{r}_t^{st}, \TargetPointSet)
     & =
    \sum_{j=1}^M
    w_j
    \Gaussian{
        \mbf{e}_{ij}(\mbf{p}_j, \mbf{m}_i, \mbf{C}_{ts}, \mbf{r}_t^{st})
        ;
        \mbf{0},
        \mbf{R}_{ij}
    },
\end{align}
with uniform weights $w_j = \frac{1}{M}$.
100 different landmark configurations are generated, with 100
different noisy point-cloud pairs generated for each. The experiments are run
for a planar 2D case where $N = 2$, and a 3D case where $N = 3$. For the 2D (3D) case, each
landmark configuration is simulated by first generating 15 (20) landmarks where each
position component is generated according to $\mathcal{U} \sim(-5, 5)$,
followed by duplicating $30\%$ of them 4 times in a Gaussian with $0.1\eye$ covariance around the original.
100 measurement pairs are then generated for each landmark.
The uncorrupted measurements are generated using the ground truth transformation
$\mbf{T}_{\TargetFrame \SourceFrame}$. The groundtruth transformation itself is
randomly sampled as $\mbf{T}_{\TargetFrame \SourceFrame} =
    \exp(\mbs{\xi}_{\TargetFrame \SourceFrame}^\wedge)$, where $(\cdot)^\wedge :
    \mathbb{R}^m \to \mathfrak{g}$, with $m = 3$ for $SE(2)$ and $m = 6$ for
$SE(3)$, and $\exp \left(\cdot
    \right) : \mathfrak{g} \to G$ is the corresponding exponential map for $SE(2)$ or $SE(3)$.
The Lie group and its corresponding Lie algebra are denoted $G$ and $\mathfrak{g}$ respectively.
In
all presented experiments, $\mbs{\xi}_{\TargetFrame \SourceFrame}^\trans=
    \begin{bmatrix}
        \mbs{\xi}^{\phi^\trans} & \mbs{\xi}^{\textrm{r}^\trans}
    \end{bmatrix}
$
with $\mbs{\xi}^{\phi} \sim \mathcal{U}(-\frac{15}{180}, \frac{15}{180})$, and $\mbs{\xi}^{\textrm{r}} \sim \mathcal{U}(-0.5, 0.5)$.
The source and target point covariances $\mbs{\Sigma}_m,\mbs{\Sigma}_f$ are each
generated randomly as $\mbs{\Sigma} = \mbf{C}\mbf{D}\mbf{C}^\trans$,
where $\mbf{D}$ is a diagonal matrix with entries generated from
$\mathcal{U}(0.1, 0.6)$ and
$\mbf{C} \in SO(N)$ is a random direction-cosine-matrix generated as
$\mbf{C} =\exp(\mbs{\xi}^\wedge)$. Each entry in $\mbs{\xi}$ is generated as
$\xi_i \sim \mathcal{U}(-\pi, \pi)$, and the $(\cdot)^\wedge$ and $\exp(\cdot)$
operators are overloaded for $SO(N)$.
Since the state is now defined on a Lie group, the Jacobian and Hessian of the
loss with respect to the state are now replaced with their Lie group
counterparts~\cite{sola2021micro}.
The results are presented in Table~\ref{tab:psr_results}.
In this case, the proposed HSM approach slightly improves ANEES.
\revision{This is expected since the Hessian approximation influences the resulting
    Laplace approximation and uncertainty characterization.}
The RMSE is similar for all
approaches,
with the exception of Max-Mixture, which has a higher RMSE, although it
converges the fastest.
%
\subsection{Lost in the Woods Dataset}
\label{sec:lost_in_the_woods}
\begin{table}
    \centering
    \caption{Average RMSE (m) on the Lost in the Woods subsequences.}
    \begin{tabular}{|*{7}{c|}} \hline
        Max Radius (m)           & \multicolumn{3}{c|}{2} & \multicolumn{3}{c|}{4}                                                                 \\ \hline
        Subseq. Len (s)          & 20                     & 30                     & 40            & 20            & 30            & 40            \\ \hline
        MM                       & \textbf{0.30}          & \textbf{0.41}          & 0.55          & \textbf{0.14} & \textbf{0.16} & \textbf{0.24} \\
        SM                       & 0.44                   & 0.51                   & 0.61          & 0.40          & 0.48          & 0.56          \\
        MSM                      & \textbf{0.30}          & \textbf{0.41}          & 0.54          & \textbf{0.14} & \textbf{0.16} & \textbf{0.24} \\
        \revision{HSM$^\dagger$} & \textbf{0.30}          & \textbf{0.41}          & \textbf{0.48} & \textbf{0.14} & \textbf{0.16} & 0.25          \\ \hline
    \end{tabular}
    \label{tab:lost_in_the_woods_errors}
    \vspace{-0.3cm}
\end{table}
\begin{table}
    \centering
    \caption{Average iterations on the Lost in the Woods subsequences.}
    \begin{tabular}{|*{7}{c|}} \hline
        Max Radius (m)           & \multicolumn{3}{c|}{2} & \multicolumn{3}{c|}{4}                                                                 \\ \hline
        Subseq. Len (s)          & 20                     & 30                     & 40            & 20            & 30            & 40            \\ \hline
        MM                       & \textbf{18.7}          & \textbf{22.3}          & 27.5          & \textbf{16.7} & \textbf{18.4} & 24.1          \\
        SM                       & 200.0                  & 200.0                  & 200.0         & 200.0         & 200.0         & 200.0         \\
        MSM                      & \textbf{18.7}          & \textbf{22.3}          & 27.6          & \textbf{16.7} & 18.8          & 24.4          \\
        \revision{HSM$^\dagger$} & \textbf{18.7}          & \textbf{22.3}          & \textbf{27.0} & \textbf{16.7} & \textbf{18.4} & \textbf{23.9} \\ \hline
    \end{tabular}
    \label{tab:lost_in_the_woods_iterations}
    \vspace{-0.5cm}
\end{table}
\begin{table}
    \centering
    \caption{\revision{Average time on the Lost in the Woods subsequences.}}
    \begin{tabular}{|*{7}{c|}}
        \hline
        Max Radius (m)  & \multicolumn{3}{c|}{2} & \multicolumn{3}{c|}{4}                                                                  \\ \hline
        Subseq. Len (s) & 20                     & 30                     & 40            & 20            & 30            & 40             \\ \hline
        MM              & \textbf{1.45}          & \textbf{3.08}          & \textbf{5.22} & \textbf{2.88} & \textbf{5.55} & \textbf{10.46} \\
        SM              & 15.04                  & 35.27                  & 49.27         & 48.86         & 87.48         & 121.64         \\
        MSM             & 1.63                   & 3.53                   & 6.22          & 3.57          & 7.16          & 13.76          \\
        HSM             & 1.60                   & 3.41                   & 5.83          & 3.38          & 6.66          & 12.76          \\ \hline
    \end{tabular}
    \label{tab:lost_in_the_woods_time}
    \vspace{-0.5cm}
\end{table}
The ``Lost in the Woods'' dataset \cite{lostInTheWoodsDataset} consists of a wheeled robot navigating a
forest of plastic tubes, which are treated as landmarks.
The robot receives wheel odometry measurements providing forward velocity and
angular velocity measurements. The robot is equipped
with a laser rangefinder that provides range-bearing measurements to the landmarks. The task is to estimate the robot poses,
$\mathcal{T} = \left\{\mathbf{T}_1^{\mathrm{rob}}, \ldots,
    \mathbf{T}_K^\mathrm{rob} \right\}$, with $\mathbf{T}_i^\mathrm{rob} \in SE(2),
    i = 1, \ldots, K$, and the
landmark positions $\mathcal{L} = \left\{\mbs{\ell}_1, \ldots, \mbs{\ell}_\NumLandmarks \right\},
    \mbs{\ell}_i \in \mathbb{R}^2, i = 1, \ldots, n_\ell$,
given the odometry and range-bearing measurements.
The process model consists of nonholonomic vehicle kinematics with the
wheel odometry forward and angular velocity inputs detailed in~\cite{cossette2020}.
The range-bearing
measurements of landmark $j$ observed from pose $k$ are in the form $\mbf{y}_{jk} = \mbf{g}
    \left(\mbf{T}_k^\mathrm{rob}, \mbs{\ell}_j \right) + \mbf{v}_{jk}$, where
$\mbf{v}_{jk} \sim \mathcal{N} \left(\mbf{0}, \mbf{R}_{jk} \right)$ and $\mbf{g}
    \left(\cdot, \cdot \right)$ is the range-bearing measurement model, also detailed in~\cite{cossette2020}. While the
dataset provides the data-association variables (i.e., which landmark
each measurement corresponds to), to evaluate the performance of the
algorithms, the challenging example of \emph{unknown data associations} is
considered. When the data associations are unknown, a multimodal measurement
likelihood may be used, as in \cite{doherty2020probabilistic}, where each measurement could have been produced from any
landmark in $\mathcal{L}$, such that
\begin{align}
    \label{eq:lost_in_the_woods_likelihood}
    p(\mbf{y}_k | \mbf{T}_k^\mathrm{rob}, \mc{L}) & =
    \sum_{j=1}^{\NumLandmarks}
    w_i \Gaussian{
        \mbf{y}_k -\mbf{g}(\mbf{T}_k^\mathrm{rob}, \mbs{\ell}_j); \mbf{0}, \mbf{R}_{jk}
    }.
\end{align}
The optimization problem to be solved then consists of process model residuals
corresponding to the nonholonomic vehicle dynamics, as well as multimodal range-bearing
error terms of the form~\eqref{eq:lost_in_the_woods_likelihood} for each
measurement. To simplify the problem, it is assumed that the groundtruth number
of landmarks are known, so that the number of components in each
mixture~\eqref{eq:lost_in_the_woods_likelihood} is known \emph{a priori}. Additionally, it is assumed that an initial guess for
each landmark position is available. This is similar to the setup of the
``oracle baseline'' described in~\cite{zhang2023}.
The dataset provides odometry and landmark measurements at 10 Hz. To reduce the problem size, the landmark
measurements are subsampled to 1 Hz.
\par
To provide an initial guess of the robot states and the landmark states, robot
poses are initialized through dead-reckoning the wheel odometry measurements,
while the landmarks are initialized by inverting the measurement model for the
first measurement of
each landmark. Note that the initialization of the landmark positions utilizes
the data association labels, but the actual optimization solves for the
associations implicitly using the multimodal likelihoods for each measurement.
\par
The dataset of length 1200 seconds is split into subsequences of length
$T_\text{subseq}$ seconds each, such that there is no overlap between
subsequences. The performance metrics computed for each subsequence
to each subsequence are averaged to provide mean metrics. For each subsequence length, two scenarios are considered that
correspond to different limits $r_{\textrm{max}}$ on the range of the
range-bearing measurements used.
\par
For each subsequence length and maximum range radius, the average RMSE and the
average number of iterations are summarized in
Table~\ref{tab:lost_in_the_woods_errors} and
Table~\ref{tab:lost_in_the_woods_iterations}, respectively.
\revision{With the exception of Sum-Mixture that fails
    due to the inacurate Hessian, performance is similar in this experiment for all methods in terms of RMSE and iteration count.
    This is explained by landmarks being
    far to each other relative to the range-bearing sensor noise characteristics.
    The less the components overlap, the smaller performance improvement can be expected from HSM relative to previous methods
    as stated in Sec.~\ref{sec:intro}.
    There is a slight improvement for runtime for HSM compared to MSM in Table~\ref{tab:lost_in_the_woods_time}.
    This difference must be interpreted with caution. Although reasonable efforts were made to optimize
    all of the approaches, the \texttt{Python} implementation may still not be representative of
    optimized real-time systems.
    Nevertheless,
    both methods require the computation of $\Jac{\rho}{f_k}, \mbf{e}_k, \Jac{\mbf{e}_k}{\mbf{x}}$ for each component to compute
    error Jacobians
    at each iteration. Thus, it is recommended to consider the iteration count as an implementation-independent
    proxy for runtime.
}
%
%
%
\FloatBarrier

\vspace{-1mm}
\section{Conclusion}
\revision{This paper proposes a novel Hessian approximation for GMMs that is compatible
  with NLS solvers. Speed and consistency improvements are 
  shown in simulation and similar performance to state of 
  the art is demonstrated in experiment. }
While aimed at improving convergence, the proposed method remains a local optimization method,
requiring a good enough initial guess
for proper operation. Future work will consider global convergence properties of the solution.
\printbibliography
\onecolumn
\appendix 
\subsection{Practical Implementation}
The Jacobians of the different mixtures as well as details of
multipliers for practical implementation are covered in this section.
The negative log likelihoods are proportional to, but not exactly equal to,
the squared error terms because of the normalization constants required to make
sure the square root argument does not become negative.
\subsubsection{Max-Mixture}
The log-likelihood is computed from the max-mixture approximation
to the Gaussian mixture such that
\begin{align}
    -\log p_{\max}(\mbf{y}|\mbf{x})
     & =
    \frac{1}{2}
    \mbf{e}_{\text{max}}^\trans \mbf{e}_{\text{max}}                          \\
     & =
    - \log \alpha_\kmax + \frac{1}{2}\mbf{e}_\kmax^\trans \mbf{e}_\kmax,
    \quad \kmax =
    \argmax \alpha_k \exp \left(-\frac{1}{2}\mbf{e}_k^\trans \mbf{e}_k\right) \\
     & \sim
    \frac{1}{2}
    \norm{
        \begin{bmatrix}
            \sqrt{2}\sqrt{\log c -\log \alpha_\kmax} \\
            \mbf{e}_\kmax
        \end{bmatrix}
    }_2^2,
\end{align}
with the normalizing constant is given by $c=\max_k \alpha_k$~\cite{pfeiferAdvancingMixtureModels2021}
to avoid the square root argument becoming negative,
and
the corresponding Jacobian given by
\begin{align}
    \frac{\del \mbf{e}_{\text{max}}}{\del \mbf{x}} & =
    \begin{bmatrix}
        \mbf{0}^{1 \times \ErrDim}
        \\ \Jacdx{\mbf{e}_\kmax}
    \end{bmatrix}.
\end{align}
\subsubsection{Sum-Mixture}
The log-likelihood is computed from the full Gaussian Mixture, and in the sum-mixture case
is given by the squared norm of a single scalar term,
\begin{align}
    -\log p_{\text{GM}}(\mbf{y}|\mbf{x}) & \sim
    \frac{1}{2}e_{\text{SM}}^2,                                                                                               \\
    \frac{1}{2}e_{\text{SM}}^2           & =
    \frac{1}{2}
    \sqrt{2\left(\log c - \log \sum_{k=1}^K\alpha_k \exp\left(-\frac{1}{2}\mbf{e}_k^\trans \mbf{e}_k\right)\right)}^2
    \label{eq:e_sum_mixture}
    \\
                                         & = \frac{1}{2}\sqrt{2\left(\log c + \frac{1}{2}\mbf{e}_\kmax^\trans \mbf{e}_\kmax -
        \log \sum_{k=1}^K \alpha_k \exp\left(-\frac{1}{2}\mbf{e}_k^\trans \mbf{e}_k+\frac{1}{2}\mbf{e}_\kmax^\trans \mbf{e}_\kmax\right)\right)}^2,
\end{align}
where the normalization constant $c$ is given by $\sum_{k=1}^K \alpha_k$,
and the Jacobian is given by
\begin{align}
    \frac{\del e_{\text{SM}}}{\del \mbf{x}} & =
    \frac{1}{2e_{\text{SM}}} \frac{-2}{\sum_{k=1}^K \alpha_k \exp\left(-\frac{1}{2}\mbf{e}_k^\trans \mbf{e}_k\right)}
    \sum_{k=1}^K\alpha_k \exp\left(-\frac{1}{2}\mbf{e}_k^\trans \mbf{e}_k\right)(-\mbf{e}_k^\trans)\Jacdx{\mbf{e}_k} \\
                                            & =
    \frac{1}{e_{\text{SM}}}
    \frac{\sum_{k=1}^K \alpha_k \exp\left(-\frac{1}{2}\mbf{e}_k^\trans \mbf{e}_k\right)\mbf{e}_k^\trans
        \frac{\del \mbf{e}_k}{\del \mbf{x}}
    }{\sum_{k=1}^K \alpha_k \exp\left(-\frac{1}{2}\mbf{e}_k^\trans \mbf{e}_k\right)}.
\end{align}
\subsubsection{Max-Sum-Mixture}
The algebraic value for the Max-Sum-Mixture is the same as for the sum-mixture, but the error
term partitioning is different, such that
\begin{align}
    -\log p_{\text{GM}}(\mbf{y}|\mbf{x}) & \sim
    \frac{1}{2}
    \mbf{e}_{\text{MSM}}^\trans \mbf{e}_{\text{MSM}} \\
                                         & =
    \frac{1}{2}
    \mbf{e}_\kmax^\trans \mbf{e}_\kmax +
    \frac{1}{2}
    \sqrt{
        2\left(
        \log c -
        \log
        \sum_{k=1}^K \alpha_k
        \exp\left(-\frac{1}{2}\mbf{e}_k^\trans \mbf{e}_k+\frac{1}{2}\mbf{e}_\kmax^\trans \mbf{e}_\kmax\right)
        \right)
    }^2
    \\
                                         & =
    \frac{1}{2}
    \norm{
        \begin{bmatrix}
            \sqrt{
                2\left(
                \log c -
                \log
                \sum_{k=1}^K \alpha_k
                \exp\left(-\frac{1}{2}\mbf{e}_k^\trans \mbf{e}_k+\frac{1}{2}\mbf{e}_\kmax^\trans \mbf{e}_\kmax\right)\right)} \\
            \mbf{e}_\kmax
        \end{bmatrix}
    }_2^2                                            \\
                                         & =
    \norm{
        \begin{bmatrix}
            e_{\text{NL}} \\
            \mbf{e}_\kmax
        \end{bmatrix}
    }_2^2,
\end{align}
where $\kmax =
    \argmax \alpha_k \exp \left(-\frac{1}{2}\mbf{e}_k^\trans \mbf{e}_k\right)$ is the dominant component
and
\begin{align}
    e_{\text{NL}} & =
    \sqrt{
        2\left(
        \log c -
        \log
        \sum_{k=1}^K \alpha_k
        \exp\left(-\frac{1}{2}\mbf{e}_k^\trans \mbf{e}_k+\frac{1}{2}\mbf{e}_\kmax^\trans \mbf{e}_\kmax\right)\right)},
    \label{eq:err_nonlinear}
\end{align}
is defined similarly to $e_{\text{SM}}$ in~\eqref{eq:e_sum_mixture}.
The error Jacobian is given by
\begin{align}
    \frac{\del \mbf{e}_{\text{MSM}}}{\del \mbf{x}} & =
    \begin{bmatrix}
        \Jacdx{\mbf{e}_\kmax} \\
        \Jacdx{e_{\text{NL}}}
    \end{bmatrix}
\end{align}
and the Jacobian of the nonlinear term $\Jacdx{e_{\text{NL}}}$ is given by
\begin{align}
    \Jacdx{e_{\text{NL}}}
     & =
    \frac{1}{e_{\text{NL}}}
    \frac{-\sum_{k=1}^K
        \alpha_k \exp
        (
        -\frac{1}{2} \mbf{e}_k^\trans \mbf{e}_k + \frac{1}{2}\mbf{e}_\kmax^\trans \mbf{e}_\kmax)
        \left(-\mbf{e}_k^\trans\frac{\del \mbf{e}_k}{\del \mbf{x}}
        + \mbf{e}_\kmax^\trans\frac{\del \mbf{e}_\kmax}{\del \mbf{x}}
        \right)
    }{\sum_{k=1}^K \alpha_k \exp\left(-\frac{1}{2}\mbf{e}_k^\trans \mbf{e}_k + \frac{1}{2}\mbf{e}_\kmax^\trans \mbf{e}_\kmax\right)}.
\end{align}
The normalization constant $c$ is given by
\begin{align}
    c=K \max \alpha_k + \delta ,
\end{align}
where $\delta$ is a damping constant~\cite{pfeiferAdvancingMixtureModels2021} that controls the influence of
the nonlinear term~\eqref{eq:err_nonlinear}.


\subsection{Robust Loss: Iterative Reweighted Least Squares}
The robust loss formulation for a single factor may be written as
\begin{align}
    J & = \rho (f(\mbf{x})) \\
      & = \rho \left(
    \frac{1}{2}\mbf{e}(\mbf{x})^\trans \mbf{e}(\mbf{x})
    \right), \label{eq:single_factor_robust_loss}
\end{align}
where $\mbf{e}: \rnums^\StateDim \rightarrow \ErrDim$ is the error function,
$\rho: \rnums \rightarrow \rnums$ is a robust loss function,
and $f(\mbf{x})=\frac{1}{2}\mbf{e}(\mbf{x})^\trans \mbf{e}(\mbf{x})$ is
a convenient intermediate quantity for the use of the chain rule.
\par
The robustified Gauss-Newton least squares update is derived by
considering Newton's method applied to~\eqref{eq:single_factor_robust_loss},
while assuming the Gauss-Newton Hessian approximation is valid for $f$,
\begin{align}
    \Hessdx{f}\approx
    \frac{\del \mbf{e}}{\del \mbf{x}^\trans}
    \Jacdx{\mbf{e}}.
    \label{eq:f_gauss_newton_approximation}
\end{align}
The Jacobian of the loss function~\eqref{eq:single_factor_robust_loss} is then given by
\begin{align}
    \Jacdx{J} & =
    \frac{\dee \rho}{\dee f}
    \Jacdx{f},
\end{align}
and the Hessian by
\begin{align}
    \Hessdx{J} & =
    \frac{\dee \rho}{\dee f}
    \Hessdx{f} +
    \frac{\del f}{\del \mbf{x}^\trans}
    \frac{\dee^2 \rho}{\dee f^2}
    \frac{\del f}{\del \mbf{x}}.
\end{align}
The intermediate function $f=\frac{1}{2}\mbf{e}^\trans \mbf{e}$
has Jacobian
\begin{align}
    \Jacdx{f} & =\mbf{e}^\trans \Jacdx{\mbf{e}},
    \label{eq:jacobian_robust_loss_approach}
\end{align}
and Hessian approximated by~\eqref{eq:f_gauss_newton_approximation}.
The Hessian of the loss is thus given by
\begin{align}
    \Hessdx{J} & =
    \frac{\dee \rho}{\dee f}
    \Hessdx{f} +
    \frac{\del f}{\del \mbf{x}^\trans}
    \frac{\dee^2 \rho}{\dee f^2}
    \frac{\del f}{\del \mbf{x}}                  \\
               & =
    \frac{\dee \rho}{\dee f}
    \frac{\del \mbf{e}}{\del \mbf{x}^\trans}
    \Jacdx{\mbf{e}} +
    \left(2\mbf{e}^\trans \Jacdx{\mbf{e}}\right)^\trans
    \frac{\dee^2 \rho}{\dee f^2}
    \left(2\mbf{e}^\trans \Jacdx{\mbf{e}}\right) \\
               & =
    \frac{\dee \rho}{\dee f}
    \frac{\del \mbf{e}}{\del \mbf{x}^\trans}
    \Jacdx{\mbf{e}} +
    \frac{\dee^2 \rho}{\dee f^2}
    \frac{\del \mbf{e}}{\del \mbf{x}^\trans}\mbf{e}
    \mbf{e}^\trans \Jacdx{\mbf{e}}.
    \label{eq:hessian_robust_loss_approx}
\end{align}
Writing the Newton method update step
with the Hessian approximation~\eqref{eq:hessian_robust_loss_approx}
and the Jacobian~\eqref{eq:jacobian_robust_loss_approach}
yields
\begin{align}
    \left(
    \frac{\dee \rho}{\dee f}
    \frac{\del \mbf{e}}{\del \mbf{x}^\trans}
    \Jacdx{\mbf{e}} +
    \frac{\dee^2 \rho}{\dee f^2}
    \frac{\del \mbf{e}}{\del \mbf{x}^\trans}\mbf{e}
    \mbf{e}^\trans \Jacdx{\mbf{e}}
    \right) \delta \mbf{x}
     & =
    -
    \frac{\dee \rho}{\dee f}
    \mbf{e}^\trans \Jacdx{\mbf{e}},
    \label{eq:hessian_robust_final}
\end{align}
Using only the robust loss function first order term
is exactly the iteratively reweighted least squares approach.
Some solvers, such as Ceres, also use the second order term, which is called the
Triggs correction.
Note that this approach does \emph{not} correspond to
definining an error term of the form
\begin{align}
    \tilde{e} & =\sqrt{\rho(\mbf{e}^\trans \mbf{e})},
\end{align}
and using that in Gauss-Newton since that would be ill-conditioned,
as the error term size reduces to one.
Furthermore, for problems with many different factors,
the robust loss $\rho$ terms will all have different values.
\subsubsection{Hessian Approximation for Sum-Mixtures}
Consider an optimization problem consisting of a single factor of the form
\begin{align}
    J & = \log c - \log \sum_{k=1^K}\alpha_k\exp
    \left(-\frac{1}{2}\mbf{e}_k^\trans \mbf{e}_k\right)                                                         \\
      & = \sqrt{\log c - \log \sum_{k=1^K}\alpha_k \exp \left(-\frac{1}{2}\mbf{e}_k^\trans \mbf{e}_k\right)}^2,
\end{align}
where the square root is present to make it similar to standard error evaluation for
Gauss-Newton, and the normalization constant $c$ is present for the square root
argument to be positive.
\par
Similarly to the robust loss case~\eqref{eq:single_factor_robust_loss},
the additional nonlinearity imposed by the Gaussian mixture needs to be taken into account
for the Hessian approximation.
Analogously to the robust loss case, $f_k=\frac{1}{2}\mbf{e}_k^\trans \mbf{e}_k$
is assumed well-behaved such that
\begin{align}
    \Hessdx{f_k}\approx
    \frac{\del \mbf{e}_k}{\del \mbf{x}^\trans}
    \Jacdx{\mbf{e}_k},
    \label{eq:f_k_gauss_newton_approximation}
\end{align}
and the definition for $\rho$ follows as
\begin{align}
    \rho(f_1, \dots, f_k) & =
    \log c - \log \sum_{k=1}^K \alpha_k\exp (-f_k).
\end{align}
Results from the robust loss case are unapplicable here due to the additional
nonlinearity due to the LogSumExp term and the presence of multiple $f_k$ terms
instead of a single one.
\par
Nevertheless, a robustified Hessian approximation for the sum-mixture may be derived.
The Jacobian is given by
\begin{align}
    \Jacdx{J} & =
    \sum_{k=1}^K
    \frac{\del \rho}{\del f_k}
    \Jacdx{f_k}, \label{eq:jac_gaussian_mixture_first}
\end{align}
and the Hessian by
\begin{align}
    \Hessdx{J} & =
    \sum_{k=1}^K
    \left(
    \frac{\del \rho}{\del f_k}
    \Hessdx{f_k} +
    \frac{\del f_k}{\del \mbf{x}^\trans}
    \sum_{j=1}^K
    \frac{\del^2 \rho}{\del f_j f_k}
    \frac{\del f_j}{\del \mbf{x}}
    \right).
    \label{eq:hessian_gaussian_mixture_first}
\end{align}
The partial derivatives of $\rho$ are given by
\begin{align}
    \frac{\del \rho}{\del f_k}            & =
    -\frac{-\alpha_k\exp(-f_k)}{\sum_{i=1}^K \alpha_i \exp(-f_i)} \\
                                          & =
    \frac{\alpha_k \exp(-f_k)}{\sum_{i=1}^K \alpha_i\exp(-f_i)}   \\
    \frac{\del^2 \rho}{\del f_j \del f_k} & =
    \frac{
        \delta_{jk} (-\alpha_k \exp(-f_k))\sum_{i=1}^K\alpha_i\exp(-f_i)
        -\alpha_k \exp(-f_k)(-\alpha_j \exp(-f_j))}
    {(\sum_{i=1}^K\alpha_i\exp(-f_i))^2}                          \\
                                          & =
    \frac{
        -\delta_{jk} (\alpha_k \exp(-f_k))\sum_{i=1}^K\alpha_i\exp(-f_i)
        +\alpha_k \alpha_j  \exp(-f_k)\exp(-f_j)}
    {(\sum_{i=1}^K\alpha_i\exp(-f_i))^2}
\end{align}
and the Jacobian of $f_k=\frac{1}{2}\mbf{e}_k^\trans \mbf{e}_k$
is given by
\begin{align}
    \Jacdx{f_k} & =\mbf{e}_k^\trans \Jacdx{\mbf{e}_k}.
\end{align}
The Jacobian of the loss $J$ is then given by~\eqref{eq:jac_gaussian_mixture_first},
\begin{align}
    \Jacdx{J} & =
    \sum_{k=1}^K
    \frac{\del \rho}{\del f_k}
    \Jacdx{f_k}                         \\
              & =
    \sum_{k=1}^K
    \frac{\alpha_k \exp(-f_k)}{\sum_{i=1}^K\alpha_i\exp(-f_i)}
    \mbf{e}_k^\trans \Jacdx{\mbf{e}_k}  \\
              & =
    \sum_{k=1}^K
    \frac{\alpha_k \exp(-f_k)}{\sum_{i=1}^K \alpha_i \exp(-f_i)}
    \mbf{e}_k^\trans \Jacdx{\mbf{e}_k}. \\
    \label{eq:jac_gaussian_mixture_second}
\end{align}
The Hessian of the loss is given by~\eqref{eq:hessian_gaussian_mixture_first}
\begin{align}
    \Hessdx{J} & =
    \sum_{k=1}^K
    \left(
    \frac{\del \rho}{\del f_k}
    \Hessdx{f_k} +
    \frac{\del f_k}{\del \mbf{x}^\trans}
    \sum_{j=1}^K
    \frac{\del^2 \rho}{\del f_j f_k}
    \frac{\del f_j}{\del \mbf{x}}
    \right)                                         \\
               & =
    \sum_{k=1}^K
    \frac{\del \rho}{\del f_k}
    \frac{\del \mbf{e}_k}{\del \mbf{x}^\trans}
    \Jacdx{\mbf{e}_k}
    +
    \left(\mbf{e}_k^\trans \Jacdx{\mbf{e}_k}\right)^\trans
    \sum_{j=1}^K
    \frac{\del^2 \rho}{\del f_j f_k}
    \left(\mbf{e}_j^\trans \Jacdx{\mbf{e}_j}\right) \\
               & =
    \sum_{k=1}^K
    \frac{\del \rho}{\del f_k}
    \frac{\del \mbf{e}_k}{\del \mbf{x}^\trans}
    \Jacdx{\mbf{e}_k}
    +
    \sum_{j=1}^K
    \frac{\del^2 \rho}{\del f_j f_k}
    \frac{\del \mbf{e}_k}{\del \mbf{x}^\trans}
    \mbf{e}_k
    \mbf{e}_j^\trans \Jacdx{\mbf{e}_j}.
\end{align}

\subsection{Derivation of Normalization Constant}
Here we derive the result related to to the lower bound on $\Delta J$ presented in
Section IV of the paper.
The solver error is given by
\begin{align}
    \mbf{e}_{\textrm{solver}}^\trans & =
    \begin{bmatrix}
        \mbf{e}_{\textrm{solver}, 1}^\trans & \sqrt{2(\GmmNormConst{\HessSumMix} + \Delta J)}
    \end{bmatrix},
\end{align}
where
\begin{align}
    \mbf{e}_{\textrm{solver}, 1}^\trans & =
    \begin{bmatrix}
        \sqrt{\frac{\del \rho}{\del f_1}}
        \mbf{e}_1^\trans & \dots &
        \sqrt{\frac{\del \rho}{\del f_K}} \mbf{e}_K^\trans
    \end{bmatrix},
\end{align}
and $\frac{\del \rho}{\del f_k}$ is defined such that
\begin{align}
    \frac{\del \rho}{\del f_k} & =
    \frac{
        \alpha_k \exp\left(-\frac{1}{2} \mbf{e}^\trans_k \mbf{e}_k \right)
    }{
        \sum_{i=1}^\NumComp \alpha_i \exp\left(-\frac{1}{2} \mbf{e}^\trans_i \mbf{e}_i \right)
    }.
\end{align}
The desired error is given by the negative log-likelihood of the GMM as
\begin{align}
    J_{\text{GMM}}(\mbf{x})
     & =
    -\log \sum_{k=1}^\NumComp
    \alpha_k
    \exp \left(-\frac{1}{2}\mbf{e}_k (\mbf{x} )^\trans \mbf{e}_k (\mbf{x} )\right).
\end{align}

By setting $\Delta J=J_{\textrm{GMM}}-\frac{1}{2}\mbf{e}_{\textrm{solver}, 1}^\trans\mbf{e}_{\textrm{solver}, 1}$,
the evaluated cost becomes
\begin{align}
    \frac{1}{2} \mbf{e}_{\textrm{solver}}^\trans  \mbf{e}_{\textrm{solver}} & =
    \frac{1}{2} \mbf{e}_{\textrm{solver},1}^\trans  \mbf{e}_{\textrm{solver},1} +
    \GmmNormConst{\HessSumMix} + \Delta J-\frac{1}{2}\mbf{e}_{\textrm{solver}, 1}^\trans\mbf{e}_{\textrm{solver}, 1}
    \\ & =
    \GmmNormConst{\HessSumMix} + J_{\textrm{GMM}}.
\end{align}
The constant $\GmmNormConst{\HessSumMix}$ is set such that $\GmmNormConst{\HessSumMix} + \Delta J \geq 0$,
which requires a lower bound on $\Delta J$.
The lower bound is obtained by first manipulating $\Delta J$ as
\begin{align}
    \Delta J & =J_{\textrm{GMM}}-\frac{1}{2}\mbf{e}_{\textrm{solver}, 1}^\trans\mbf{e}_{\textrm{solver}, 1} \\
             & =
    -\log \sum_{k=1}^\NumComp
    \alpha_k
    \exp \left(-\frac{1}{2}\mbf{e}_k (\mbf{x} )^\trans \mbf{e}_k (\mbf{x} )\right)
    -\frac{1}{2}\mbf{e}_{\textrm{solver}, 1}^\trans\mbf{e}_{\textrm{solver}, 1}                             \\
             & =
    -\log \sum_{k=1}^\NumComp
    \alpha_k
    \exp \left(-\frac{1}{2}\mbf{e}_k (\mbf{x} )^\trans \mbf{e}_k (\mbf{x} )\right)
    -\frac{1}{2}
    \sum_{k=1}^\NumComp
    \frac{
        \alpha_k \exp\left(-\frac{1}{2} \mbf{e}^\trans_k \mbf{e}_k \right)
    }{
        \sum_{i=1}^\NumComp \alpha_i \exp\left(-\frac{1}{2} \mbf{e}^\trans_i \mbf{e}_i \right)
    }
    \mbf{e}_k^\trans \mbf{e}_k.
\end{align}
Changing variables to $f_k = \frac{1}{2}\mbf{e}_k (\mbf{x} )^\trans \mbf{e}_k (\mbf{x})$ yields
\begin{align}
    \Delta J & =
    -\log \sum_{k=1}^\NumComp
    \alpha_k
    \exp \left(-f_k  \right)
    -\frac{1}{2}
    \sum_{k=1}^\NumComp
    \frac{
        \alpha_k \exp\left(-f_k  \right)
    }{
        \sum_{i=1}^\NumComp \alpha_i \exp\left(-f_i \right)
    }
    2f_k         \\
             & =
    -\log \sum_{k=1}^\NumComp
    \alpha_k
    \exp \left(-f_k  \right)
    -
    \sum_{k=1}^\NumComp
    \frac{
        \alpha_k \exp\left(-f_k  \right)
    }{
        \sum_{i=1}^\NumComp \alpha_i \exp\left(-f_i \right)
    }
    f_k,
\end{align}
with $\alpha_k>0, f_k\geq 0$.
Then, exponentiating and taking logarithm of the second term yields
\begin{align}
    \Delta J & =
    -\log \sum_{k=1}^\NumComp
    \alpha_k
    \exp \left(-f_k  \right)
    -
    \sum_{k=1}^\NumComp
    \frac{
        \alpha_k \exp\left(-f_k  \right)
    }{
        \sum_{i=1}^\NumComp \alpha_i \exp\left(-f_i \right)
    }
    f_k          \\
             & =
    -\log
    \left(
    \sum_{k=1}^\NumComp
    \alpha_k
    \exp \left(-f_k  \right)
    \right)
    -\log
    \left(
    \exp
    \sum_{k=1}^\NumComp
    \frac{
        \alpha_k \exp\left(-f_k  \right)
    }{
        \sum_{i=1}^\NumComp \alpha_i \exp\left(-f_i \right)
    }
    f_k
    \right)
    \\
             & =
    -\log \sum_{k=1}^\NumComp
    \left(
    \alpha_k
    \exp \left(-f_k
    \right)
    \exp
    \sum_{j=1}^\NumComp
    \frac{
        \alpha_j \exp\left(-f_j  \right)
    }{
        \sum_{i=1}^\NumComp \alpha_i \exp\left(-f_i \right)
    }
    f_j
    \right)
    \\ &=
    -\log \sum_{k=1}^\NumComp
    \left(
    \alpha_k
    \exp \left(-f_k+
        \sum_{j=1}^\NumComp
        \frac{
            \alpha_j \exp\left(-f_j  \right)
        }{
            \sum_{i=1}^\NumComp \alpha_i \exp\left(-f_i \right)
        }
        f_j
        \right)
    \right)      \\
             & =
    -\log \sum_{k=1}^\NumComp
    \left(
    \alpha_k
    \exp \left(
        \sum_{j=1}^\NumComp
        \frac{
            \alpha_j \exp\left(-f_j  \right)
        }{
            \sum_{i=1}^\NumComp \alpha_i \exp\left(-f_i \right)
        }
        \left(f_j-f_k\right)
        \right)
    \right).
    \label{eq:delta_J_manipulated}
\end{align}
The overall expression in~\eqref{eq:delta_J_manipulated} is decreasing as a function of
the exponent argument $S_k$,
\begin{align}
    S_k & =
    \sum_{j=1}^\NumComp
    \frac{
        \alpha_j \exp\left(-f_j  \right)
    }{
        \sum_{i=1}^\NumComp \alpha_i \exp\left(-f_i \right)
    }
    \left(f_j-f_k\right).
\end{align}
Therefore, an upper bound on $S_k$ is required.
In the worst case scenario, $f_j \geq f_k$ for all $j$ in the summation.
This can be clarified as follows.
Defining $A_{f_j \geq f_k}$ as the set of indices $j$ with $f_j\geq f_k$,
as well as the set $A_{f_j < f_k}$ with $f_j < f_k$ with $f_j< f_k$,
and
splitting $S_k$ up into sums over these two sets yields
\begin{align}
    S_k & =
    \sum_{A_{f_j \geq f_k}}
    \frac{
        \alpha_j \exp\left(-f_j  \right)
    }{
        \sum_{i=1}^\NumComp \alpha_i \exp\left(-f_i \right)
    }
    \left(f_j-f_k\right) +
    \sum_{A_{f_j < f_k}}
    \frac{
        \alpha_j \exp\left(-f_j  \right)
    }{
        \sum_{i=1}^\NumComp \alpha_i \exp\left(-f_i \right)
    }
    \left(f_j-f_k\right)
    \label{eq:S_k_split_into_two_intermediary}
    \\
        & \leq
    \sum_{A_{f_j \geq f_k}}
    \frac{
        \alpha_j \exp\left(-f_j  \right)
    }{
        \sum_{i=1}^\NumComp \alpha_i \exp\left(-f_i \right)
    }
    \left(f_j-f_k\right),
\end{align}
since $f_j-f_k<0$ in the second term of~\eqref{eq:S_k_split_into_two_intermediary}.
Thus, proceeding with $f_j-f_k\geq 0$ allows to write
\begin{align}
    S_k & =
    \sum_{j=1}^\NumComp
    \frac{
        \alpha_j \exp\left(-f_j  \right)
    }{
        \sum_{i=1}^\NumComp \alpha_i \exp\left(-f_i \right)
    }
    \left(f_j-f_k\right) \\
        & \leq
    \sum_{j=1}^\NumComp
    \frac{
        \alpha_j \exp\left(-f_j  \right)
    }{
        \alpha_j \exp\left(-f_j  \right) + \alpha_k \exp\left(-f_k  \right)
    }
    \left(f_j-f_k\right) \\
        & =
    \sum_{j=1}^\NumComp
    \frac{
        1
    }{
        1 + \frac{\alpha_k}{\alpha_j} \exp\left(f_j-f_k  \right)
    }
    \left(f_j-f_k\right)
    \\
        &
    \leq
    \sum_{j=1}^\NumComp
    \frac{
        1
    }{
        \frac{\alpha_k}{\alpha_j} \exp\left(f_j-f_k  \right)
    }
    \left(f_j-f_k\right) \\
        & =
    \frac{1}{\alpha_k}
    \sum_{j=1}^\NumComp
    \alpha_j \exp\left(-(f_j-f_k)  \right) \left(f_j-f_k\right).
\end{align}
Using the change of variables $t=f_j-f_k$, with $t>0$, makes explicit that
\begin{align}
    \exp \left(-(f_j-f_k)  \right) \left(f_j-f_k\right) & =
    \exp \left(-t  \right) \left(t\right),
\end{align}
which has a single maximum at $t=1$ with $\exp \left(-t  \right) \left(t\right) = \exp(-1)$.
Therefore,
\begin{align}
    S_k & \leq
    \frac{1}{\alpha_k}
    \sum_{j=1}^\NumComp
    \alpha_j \exp(-1)
    \\
        & \leq
    \frac{1}{\alpha_k}
    \sum_{j=1}^\NumComp
    \alpha_j.
\end{align}
Therefore,
\begin{align}
    \Delta J
     & =
    -\log \sum_{k=1}^\NumComp
    \left(
    \alpha_k
    \exp \left(
        S_k
        \right)
    \right)
    \geq
    -\log \sum_{k=1}^\NumComp
    \left(
    \alpha_k
    \exp \left(
        \frac{1}{\alpha_k}
        \sum_{j=1}^\NumComp \alpha_j
        \right)
    \right),
\end{align}
and the normalization constant $\GmmNormConst{\HessSumMix}$ may be set to
\begin{align}
    \GmmNormConst{\HessSumMix} & =
    \log \sum_{k=1}^\NumComp
    \left(
    \alpha_k
    \exp \left(
        \frac{1}{\alpha_k}
        \sum_{j=1}^\NumComp \alpha_j
        \right)
    \right).
\end{align}
\subsection{Identities}
\subsubsection{Jacobian of vector-scalar product}
For $a(\mbf{x})$ a scalar function of $\mbf{x}$
and $\mbf{v}(\mbf{x})$ a column
vector function of $\mbf{x}$, the Jacobian of their product
is obtained as
\begin{align}
    \left.\Jacdx{a\mbf{v}}\right|_{i,j} & =
    \frac{\del}{\del x_j}
    a v_i                                   \\
                                        & =
    \frac{\del a}{\del x_j} v_i +
    a \frac{\del v_i}{\del x_j}             \\
    \Jacdx{a\mbf{v}}                    & =
    \mbf{v} \Jacdx{a}  + a \Jacdx{\mbf{v}}
\end{align}
\subsubsection{Hessian of a function through Jacobian of gradient}
The Hessian $\Hessdx{J}$ of a function $J(x)$ is given by a matrix with entries
\begin{align}
    \left. \Hessdx{J} \right|_{i, j} & =
    \frac{\del J}{\del x_i \del x_j}.
\end{align}
The gradient $\Jacdx{J}^\trans$ of $J$ has entries given by
\begin{align}
    \left.\Jacdx{J}^\trans \right|_{i} & =\frac{\del J}{\del x_i},
\end{align}
and the Jacobian of the gradient is given by
\begin{align}
    \left.\frac{\del}{\del \mbf{x}}\Jacdx{J}^\trans \right|_{i,j} & =
    \frac{\del J}{\del x_j \del x_i}.
\end{align}
Therefore,
\begin{align}
    \Hessdx{J} & = \left(\frac{\del}{\del \mbf{x}}\Jacdx{J}^\trans\right)^\trans.
\end{align}

\end{document}